\documentclass[twoside]{article}

\usepackage[accepted]{aistats2021}
\usepackage{mydefs}
\usepackage{algorithm}
\usepackage{algorithmic}

\hypersetup{
    colorlinks,
    linkcolor={red!90!black},
    citecolor={blue!50!black},
    urlcolor={blue!80!black}
}

\usepackage[round]{natbib}

\newcommand{\ourtitle}{Causal Autoregressive Flows}
\newcommand{\carefl}{CAREFL }
\newcommand{\carefle}{CAREFL}

\begin{document}

\twocolumn[

\aistatstitle{\ourtitle}
\aistatsauthor{ Ilyes Khemakhem$^*$ \And Ricardo P. Monti$^*$ \And  Robert Leech \And Aapo Hyv\"arinen }
\aistatsaddress{Gatsby Unit, UCL \And Gatsby Unit, UCL \And King's College London \And University of Helsinki } ]

\begin{abstract}
Two apparently unrelated fields --- normalizing flows and causality --- have recently received considerable attention in the machine learning community.
In this work, we highlight an intrinsic correspondence between a simple family of autoregressive normalizing flows and identifiable causal models.
We exploit the fact that autoregressive flow architectures define an ordering over variables, analogous to a causal ordering, to show that they are well-suited to performing a range of causal inference tasks, ranging from causal discovery to making interventional and counterfactual predictions.
First, we show that causal models derived from both affine and additive autoregressive flows
with fixed orderings over variables
are identifiable, i.e.\ the true direction of causal influence can be recovered.
This provides a generalization of the additive noise model well-known in causal discovery.
Second, we derive a bivariate measure of causal direction based on likelihood ratios, leveraging the fact that flow models can estimate normalized log-densities of data.
Third, we demonstrate that flows naturally allow for direct evaluation of both interventional and counterfactual queries, the latter case being possible due to  the invertible nature of flows.
Finally, throughout a series of experiments on synthetic and real data, the proposed method is shown to outperform current approaches for causal discovery as well as making accurate interventional and counterfactual predictions.
\end{abstract}

\section{INTRODUCTION}
\label{sec:intro}

Causal models play a fundamental role in modern scientific endeavour \citep{spirtes2000causation, pearl2009causality}.
Many of the questions which drive scientific research are not associational but rather causal in nature.
While randomized controlled studies are the gold standard for understanding the underlying causal mechanisms of a system,
such experiments are often unethical, too expensive, or technically impossible.
In the absence of randomized controlled trials, the framework of \textit{structural equation models} (SEMs) can be used to encapsulate causal knowledge as well as to answer interventional and counterfactual queries \citep{pearl2009causal}.
At a fundamental level, SEMs define a generative model for data based on causal relationships,
and contain strictly more information than their corresponding causal graph and law.

The first step in performing causal inference is to determine the
underlying causal graph. Whilst this can be achieved in several ways (e.g., randomized study, expert judgement), data driven approaches using purely observational data, termed \textit{causal discovery},
are often employed.
The challenge for causal discovery algorithms is that
given a (typically empirical) data distribution
one
can write many different SEMs that could generate such distribution
\citep{zhang2015estimation,spirtes2016causal}.
In other words, the causal structure is unidentifiable in the absence of any constraints.

Causal discovery algorithms typically take one of two approaches to achieve identifiability. The first approach is to introduce constraints over the family of
functions present in the SEM, for example assuming all causal
dependencies are linear or that disturbances are additive
\citep{shimizu2006linear,shimizu2011directlingam,hoyer2009nonlinear,peters2014causal,bloebaum2018causeeffect,zheng2018dags}. While such
approaches have been subsequently extended to allow for
bijective transformations \citep{zhang2009identifiability},
they often introduce unverifiable assumptions over the true underlying SEM.
An alternative approach is to consider
unconstrained causal models whilst introducing further
assumptions over the data distribution.
These methods often introduce non-stationarity constraints on
the distribution of latent variables \citep{peters2016causal,monti2019causal} or assume exogeneous variables are present \citep{zhang2017causal}.

In the present contribution,
we consider the first approach, i.e.\ constraining the functions defining the causal relationships,
and combine it with the framework of \textit{normalizing flows} recently developed in deep learning literature.

Normalizing flows \citep{papamakarios2019normalizing,kobyzev2020normalizing} provide a general way of constructing flexible generative models with tractable distributions,
where both sampling and density estimation are efficient and exact.
Flows model the data as an invertible transformation of some noise variable, whose distribution is often chosen to be simple, and make use of the change of variable formula in order to express the data density.
This formula requires the evaluation of the Jacobian determinant of the transformation.

Autoregressive normalizing flows \citep{kingma2016improving,papamakarios2018masked,huang2018neural} purposefully yield a triangular Jacobian matrix, and the Jacobian determinant can be computed in linear time.
Importantly for our purposes, the autoregressive structure in such flows is specified by an ordering on the input variables, and each output variable is only a function of the input variables that precede it in the ordering.
Different architectures for autoregressive flows have been proposed, ranging from simple additive and affine transformations \citep{dinh2014nice,dinh2016density}, to more complex cubic and neural spline transformations \citep{durkan2019cubicspline,durkan2019neural}.
Flows have been increasingly popular, with applications in density estimation \cite{dinh2016density,papamakarios2018masked}, variational inference \citep{rezende2015variational,kingma2016improving} and image generation \citep{kingma2018glow,durkan2019neural}, to name a few.
Active research is conducted in order to increase the expressivity and flexibility of flow models, while maintaining the invertibility and sampling efficiency.

In this work,
we consider the ordering of variables in an autoregressive flow model from a causal perspective,
and highlight the similarities between SEMs and autoregressive flows.
We show that under some constraints, autoregressive flow models are well suited to performing a variety of causal inference tasks.
As a first contribution, we focus on the class of affine normalizing flows, and show that it defines an identifiable causal model.
This causal model is a new generalization of the well-known additive noise model, and the proof of its identifiability constitutes the main theoretical result of this manuscript.
We then leverage the properties of flows to perform causal discovery and inference in such a models.
First, we use the fact that flows can efficiently evaluate exact likelihoods to propose a non-linear measure of causal direction based on likelihood ratios, with ensuing optimality properties.
Second, we show that when autoregressive flow models are conditioned upon the correct causal ordering, they can be  employed to accurately answer interventional and counterfactual queries.
Finally, we show that our method performs favourably on a range of experiments, both on synthetic and real data, when compared to previous methods.

\section{PRELIMINARIES}
\label{sec::background}

\subsection{Structural Equation Models}
\label{sec::SEM}

Suppose we observe $d$-dimensional random variables $\textbf{x}= (x_1, \ldots, x_d)$
with joint distribution $\mathbb{P}_{\xb}$.
A structural equation model (SEM) is here defined as a tuple $\Scal = (\Sb, \PP_\nb)$ of a collection $\Sb$ of $d$ structural equations:
\begin{equation}
\label{SEM_eq}
S_j:\quad x_j = f_j ( \textbf{pa}_j, n_j), ~~  j=1, \ldots ,d
\end{equation}
together with a joint distribution, $\mathbb{P}_\textbf{n}$, %
over latent disturbance (noise) variables, $n_j$, which are assumed to be mutually independent.
We write $\textbf{pa}_j$ to denote the parents of the variable $x_j$.
The SEM defines the \emph{observational} distribution of the random vector $\mathbf{x}$: sampling from $\mathbb{P}_\mathbf{x}$ is equivalent to sampling from $\mathbb{P}_\nb$ and propagating the samples through $\mathbf{S}$.
The causal graph $\mathcal{G}$, associated with an SEM \eqref{SEM_eq} is a graph consisting of one node corresponding to each variable $x_j$;
throughout this work we assume $\mathcal{G}$ is a directed acyclic graph (DAG).

It is well known that for a DAG, there exists a causal ordering (or permutation) $\pi$ of the nodes, such that $\pi(i) < \pi(j)$ if the variable $x_i$ precedes the variable $x_j$ in the DAG (but such an ordering is not necessarily unique).
Thus, given the causal ordering of the associated DAG we may re-write equation \eqref{SEM_eq} as
\begin{equation}
\label{SEM_eq_causalOrder}
x_j = f_j \left (  {\xb}_{ < \pi(j)}  ,  n_j \right ), ~~ ~ j=1, \ldots ,d
\end{equation}
where $\mathbf{x}_{< \pi(j)} = \{x_i: \pi(i) < \pi(j)\}$ denotes all variables before $x_j$ in the causal ordering.
Moreover, in the above definition of SEMs we allow $f_j$ to be any (possibly non-linear) function.
\cite{zhang2015distinguishing} proved that the causal direction of the general SEM~\eqref{SEM_eq} is not identifiable without constraints.
To this end, the causal discovery community has focused on specific special cases in order to obtain identifiability results as well as provide practical algorithms.
In particular, the additive noise model \citep[ANM]{hoyer2009nonlinear}, which assumes the noise is additive,
is of interest to us in the rest of this manuscript, and its SEM has the form
\begin{equation}
\label{eq:anm}
x_j = f_j ( \xb_{<\pi(j)}) + n_j, ~~ ~ j=1, \ldots ,d
\end{equation}

\subsection{Autoregressive Normalizing Flows}

Normalizing flow models seek to express the log-density of observations $\textbf{x}\in\mathbb{R}^d$ as an invertible and differentiable transformation $\Tb$ of latent variables,
$\mathbf{z}\in\mathbb{R}^d$,
which follow a simple (typically factorial) base distribution that has density $p_{\zb}(\mathbf{z})$.
This allows for the density of $\mathbf{x}$ to be obtained via a change of variables as follows:
\begin{equation*}
p_\xb(\mathbf{x}) = p_{\zb}(\Tb^{-1}(\xb)) \snorm{\det J_{\Tb^{-1}}(\mathbf{x})}
\label{flow_inverse}
\end{equation*}
Typically,
$\Tb$ or $\Tb^{-1}$
will be implemented with neural networks.
Very often, normalizing flow models are obtained by chaining together different transformations $\Tb_1, \dots, \Tb_k$ from the same family to obtain $\Tb = \Tb_1\circ\cdots\circ\Tb_k$, while remaining invertible and differentiable. The Jacobian determinant of $\Tb$ can simply be computed from the Jacobian determinants of the sub-transformations $\Tb_l$.
As such, an important consideration is ensuring the Jacobian determinant of each of the sub-transformations to be efficiently calculated.

\textit{Autoregressive} flows use transformations that are designed precisely to enable simple Jacobian computation by restricting their Jacobian matrices to be lower triangular \citep{huang2018neural}.
In this case, the transformation $\Tb$ has the form:
\begin{equation}
\label{eq:ar_flow}
x_j = \tau_j(z_j, \xb_{<\pi(j)})
\end{equation}
where $\pi$ is a permutation that specifies an autoregressive structure on $\xb$ and the functions $\tau_j$ (called \emph{transformers}) are invertible with respect to their first arguments and are parametrized by their second argument.

\section{CAUSAL AUTOREGRESSIVE FLOW MODEL}
\label{sec:model}

The ideas presented in this manuscript highlight the similarities between equations~\eqref{SEM_eq_causalOrder} and~\eqref{eq:ar_flow}.
In particular, both models explicitly define an ordering over variables and both models assume the latent variables (denoted by $\textbf{n}$ or $\textbf{z}$ respectively) follow simple, factorial distributions.
Throughout the remainder of this paper, we will look to build upon these similarities in order to employ
autoregressive flow models for causal inference.
First, we explicit in Section~\ref{sec:semflows} the general conditions under which such correspondence is possible.
Then,
we consider bivariate \emph{affine} flows in Section~\ref{sec:model_def},
and %
show that they define a causal model which is identifiable, and which generalizes existing models, in particular additive noise models.
In Section~\ref{sec::flowCD}, we present our measure of causal direction based on the ratio of the likelihoods under two alternative flow models corresponding to different causal orderings.
Finally, Section~\ref{sec:extensions} presents an extension to the multivariate case.
The causal model as well as the flow-based likelihood ratio measure of causal direction constitute the
\textbf{c}ausal \textbf{a}uto\textbf{re}gressive \textbf{fl}ow (\carefle) model.

\subsection{From Autoregressive Flow models to SEMs}
\label{sec:semflows}
There are some constraints we need to make on how we define autoregressive normalizing flows so that they remain compatible
with causal models:
\begin{enumerate}[label=({\bf\Roman*})]
	\item \label{ass1} \textbf{Fixed ordering:} When chaining together different autoregressive transformations $\Tb_1, \dots, \Tb_k$ into $\Tb = \Tb_1\circ\cdots\circ\Tb_k$, the ordering $\pi$ of the input variables should be the same for all sub-transformations.
	\item \label{ass2} \textbf{Affine/additive transformations:} The transformers $\tau_j$ in~\eqref{eq:ar_flow} take what is called an \emph{affine} form: %
	\begin{equation}
	\label{eq:affine_tau}
		\tau_j(u, \vb) = e^{s_j(\vb)}u + t_j(\vb)
	\end{equation} %
	where an \emph{additive} transformation is a special case with $s_j = 0$.
\end{enumerate}

Constraint~\ref{ass1} ensures that composing transformations maintains the autoregressive structure of the flow, so as
to respect the correspondence with a
SEM~\eqref{SEM_eq_causalOrder}.
In fact, if all sub-transformations $\Tb_l$ are autoregressive and follow the same ordering $\pi$, then $\Tb$ is also autoregressive and follows $\pi$ (see Appendix~\ref{app:transitive} for a proof).
We emphasize this point here because it is contrary to the common practice of changing the ordering $\pi$ throughout the flow to make sure all input variables interact with each other \citep{germain2015made,dinh2016density,kingma2018glow}.

Constraint~\ref{ass2} ensures that the flow model is not too flexible, and in particular cannot approximate any density.
In fact, the causal ordering of autoregressive flows with universal approximation capability is not identifiable.
A proof can be found using the theory of non-linear ICA
\citep{hyvarinen1999nonlinear}:
we can autoregressively and \emph{in any order} transform any random vector into independent components with simple distributions.
In other words, for any two variables $x_1$ and $x_2$, we can construct another variable $z_2$ such that $z_2 \independent x_1$.
Such construction is invertible for $x_2$, meaning that we can write $x_2$ as a function of $(x_1, z_2)$.
Similarly, the same treatment can be done in the reverse order, to construct a variable $z_1$ that is independent of $x_2$, such that $x_1$ is a function of $(x_2, z_1)$.
That is, any two variables would be symmetric according to the SEM.
This is in contradiction with the definition of identifiability of a causal model,
which states that the transformation $\Tb$ from noise $\zb$ to observed variable $\xb$ has a unique causal ordering.
Fortunately,
flows based on \emph{additive} and \emph{affine} transformations, as defined above (based on \citet{dinh2016density}), are not universal density approximators (see Appendix~\ref{app:ununiversality} for a proof).

Finally, note that constraints~\ref{ass1} and~\ref{ass2} only limit the expressivity of flows as universal \emph{density} approximators.
In contrast, the coefficients $s_j$ and $t_j$ of the affine transformer~\eqref{eq:affine_tau}, when parametrized as neural networks, can be universal \emph{function} approximators. This property of universal approximation of the functional relationships is preserved when stacking flows (see Appendix~\ref{app:universal} for a proof).

\subsection{Model Definition and Identifiability}
\label{sec:model_def}

Suppose we observe bivariate data $\xb = (x_1, x_2) \in \RR^2$.
Underlying the data, there is a causal ordering described by a permutation $\pi$ of the set $\{1, 2\}$, where $\pi=(1, 2)$ if $x_1\rightarrow x_2$ and $\pi=(2, 1)$ otherwise.

As per Constraints~\ref{ass1} and~\ref{ass2}, let $\Tb_1, \dots, \Tb_k$ be $k\geq1$ \emph{affine} autoregressive transformations---\ie of the form~\eqref{eq:ar_flow} where the transformers $\tau_j$ are affine functions~\eqref{eq:affine_tau}---with ordering $\pi$,
and let $\Tb=\Tb_1\circ\cdots\circ\Tb_k$.
Then $\Tb$ is also an affine transformation (see Appendix~\ref{app:transitive} for a proof).
As mentioned earlier, such \textit{composability} is a central and well-known property of affine flows: the ordering stays the same and the composition is still an affine flow.

The flow $\Tb$ defines the following SEM on the observations $\xb$:
\begin{equation}
\label{eq:flow_def_x_ns}
    x_j = e^{s_j(x_{<\pi(j)})}z_j + t_j(x_{<\pi(j)}),\quad j=1,2
\end{equation}
where $z_1,z_2$ are statistically independent latent noise variables, and $s_j(x_{<\pi(j)})$ and $t_j(x_{<\pi(j)})$ are defined constant (with respect to $\xb$) for $\pi(j) = 1$.
Equation~\eqref{eq:flow_def_x_ns} defines our proposed causal model where the noise is not merely added to some function of the cause (as typical in existing models), but also modulated by the cause.

As a special case, if the transformations $\Tb_l,\, l=1,\dots,k$ are additive (in the sense defined above), then the flow $\Tb$ is also additive, and $s_1 = s_2 = 0$.
In such a special case, Equation~\eqref{eq:flow_def_x_ns} is part of the additive noise model family~\eqref{eq:anm}, which was proven to be identifiable by~\cite{hoyer2009nonlinear}.

We present next a non-technical Theorem which states that the more general affine causal model~\eqref{eq:flow_def_x_ns} is also identifiable, when
the noise variable $\zb$ is Gaussian.
A more rigorous treatment as well as the proof of a more general case
can be found in Appendix~\ref{app:iden}.

\begin{theorem}[Identifiability]
\label{th:iden}
Assume $\xb=(x_1,x_2)$ follows the model described by equation~\eqref{eq:flow_def_x_ns}, with $z_1,z_2$ statistically independent, and the function $t_j$ linking cause to effect is non-linear and invertible.
If $z_1$ and $z_2$ are Gaussian, the model is identifiable (i.e., $\pi$ is uniquely defined).
Alternatively \citep{hoyer2009nonlinear}, if $s_1=s_2=0$, the model is identifiable for any (factorial) distribution of the noise variables $z_1$ and $z_2$.
\end{theorem}

Note that while the main result in Theorem~\ref{th:iden} assumes Gaussian noise,
we believe that the identifiability result also holds for general noise.
We show that empirically in Section~\ref{sec:exp}.

\subsection{Choosing Causal Direction using Likelihood Ratio}
\label{sec::flowCD}

Next, we use our flow-based framework to develop a concrete method for estimating the causal direction, i.e.\ $\pi$. We follow \citet{hyvarinen2013pairwise} and pose causal discovery as a statistical testing problem which we solve by likelihood ratio testing. We seek to compare two candidate models which can be seen as corresponding to two hypotheses: $x_1 \rightarrow x_2$ against $x_1 \leftarrow x_2$.
Likelihood ratios are, in general, an attractive way to deciding between alternative hypotheses (models)
because they have been proven to be uniformly most powerful, at least when testing "simple" hypotheses \citep{neyman1933ix}. However, in our special case, the framework in fact reduces to simply choosing the causal direction which has a higher likelihood.

Normalizing flows allow for easy and exact evaluation of the likelihoods.
If we assume the causal ordering $\pi=(1,2)$, then the likelihood of an affine autoregressive flow is:
\begin{multline*}
\log L_{ \pi=(1,2) } (\mathbf{x}) =
		\log p_{z_1} \left(e^{-s_1}(x_1 - t_1)\right) \\
		+ \log p_{z_2} \left(e^{-s_2(x_1)}(x_2 - t_2(x_1))\right)
		- s_1 - s_2(x_1)
\end{multline*}
We propose to fit two affine autoregressive flow models~\eqref{eq:flow_def_x_ns}, each conditioned on a
distinct causal order over variables: $\pi = (1,2)$ or $\pi=(2,1)$.
For each candidate model we train parameters for each flow via maximum likelihood.
In order to avoid overfitting we look to evaluate log-likelihood for each model over a
held out testing dataset. As such, the proposed measure of causal direction is
defined as:
\begin{multline}
\label{eq:flowLR}
 R = \EE\left[\log {L_{\pi=(1,2) }(\mathbf{x}_{test}; \mathbf{x}_{train})}\right] \\- \EE\left[\log { L_{\pi=(2,1) }(\mathbf{x}_{test}; \mathbf{x}_{train}) }\right]
\end{multline}
where $\EE\left[\log L_{\pi=(1,2)}(\mathbf{x}_{test}; \mathbf{x}_{train})\right] $ is the empirical expectation of the estimated log-likelihood
evaluated on unseen test data $\mathbf{x}_{test}$. %
If $R$ is positive we conclude that $x_1$ is the causal variable and if $R$ is negative we conclude that $x_2$ is the causal variable.

\subsection{Extension to Multivariate Data}
\label{sec:extensions}
We can generalize the likelihood ratio measure developed in section~\ref{sec::flowCD} to the multivariate case by computing the log-likelihood $\log L_\pi$ for each ordering $\pi$, and accept the ordering with highest log-likelihood as the true causal ordering of the data.
This procedure is only feasible for small values of $d$, since the numbers of permutations of $\iset{1, d}$ grows exponentially with $d$.
An alternative approach is to employ the bivariate likelihood ratio~\eqref{eq:flowLR} in conjunction with a traditional constraint based method such as the PC algorithm, similarly to \cite{zhang2009identifiability}. The PC algorithm is first used to estimate the skeleton of the DAG $G$ that describes the causal structure of the data, and orient as many edges as possible. Then, the remaining edges are oriented using the likelihood ratio measure.

We can also extend the likelihood ratio measure in a different way: we can identify the causal direction between pairs of multivariate variables.
More specifically, consider two random vectors $(\xb_1, \xb_2) \in \RR^{2d}$, and suppose that $\xb_1 \rightarrow \xb_2$.
Then they can be described by the following SEM:
\begin{equation*}
\begin{aligned}
    \xb_1 &= e^{\sb_1}\cdot\zb_1 + \tb_1 \\
    \xb_2 &= e^{\sb_2(\xb_1)}\cdot\zb_2 + \tb_2(\xb_1)
\end{aligned}
\end{equation*}
where $(\zb_1, \zb_2)$ is the vector of latent noise variables that are supposed independent, $\sb_i$ and $\tb_i$ are vector-valued instead of scalar-valued, and $\cdot$ denotes the elementwise product.
The likelihood ratio measure~\eqref{eq:flowLR} can be used straightforwardly here to find the correct causal direction between $\xb_1$ and $\xb_2$.
Note that while the identifiability theory was developed for the bivariate case, our experiments in Section~\ref{sec:exp} show that it also holds for this case of two multivariate $\xb_i$.
To the best of our knowledge, this is the first model that can readily perform causal discovery over groups of multivariate variables.

\section{CAUSAL INFERENCE USING AUTOREGRESSIVE FLOWS}
\label{sec::flowCI}

In this section we demonstrate how flow architectures may be employed to perform both interventional and counterfactual queries.
We assume that the true causal ordering over variables has been resolved (e.g.,  as the result of expert judgement or obtained via the method described in Section~\ref{sec:model}).
Interventional queries involve marginalization over latent variables and thus can be evaluated by propagating forward the structural equations.
However, counterfactual queries require us to condition, as opposed to marginalize, over latent variables.
This requires us to first infer the posterior distribution of latent variables,
termed \textit{abduction} by \citet{pearl2009causal}.
In many causal inference models this is challenging, often requiring complex inference algorithms.
However, the invertible nature of flows means that the posterior of latent variables given observations can be readily obtained.

\subsection{Interventions}
\label{sec:intervention}

It is possible to manipulate an SEM $\mathcal{S}$ to create interventional distributions over $\textbf{x}$.
As described in \citet{pearl2009causality}, intervention on a given variable $x_i$ defines a new \textit{mutilated} generative model where the structural equation associated with variable $x_i$ is replaced by the interventional value,
while keeping the rest of the equations~\eqref{eq:ar_flow} fixed.
Interventions are very useful in understanding causal relationships. If,
under the assumption of
faithfulness,
intervening on a variable $x_i$ changes the marginal distribution of another variable $x_j$, then it is likely that $x_i$ has some causal effect on $x_j$. Conversely, if intervening on $x_j$ doesn't change the marginal distribution of $x_i$, then the latter is not a descendant of $x_j$.
We follow \citet{pearl2009causality} and denote by $do(x_i=\alpha)$ the interventions that puts a point mass on $x_i$.

Autoregressive flow modelling allows us to answer interventional queries easily.
After fitting a flow model~\eqref{eq:ar_flow} conditioned on the right causal ordering (assumed known) to the data,
we change the structural equation for variable $x_i$ from $x_i = \tau_i(z_i, \xb_{<\pi(i)})$ to $x_i = \alpha$.
This breaks the edges from $x_{<\pi(i)}$ to $x_i$, and puts a point mass on the latent variable $z_i$.
Thereafter, we can directly draw samples from the distribution $\prod_{j\neq i}p_{z_j}$ for all remaining latent variables $z_{j \neq i}$.
Finally, we obtain a sample for $\xb^{do(x_i=\alpha)}$ by passing these samples through the flow,
which allows us to compute empirical estimates of the interventional distribution.
This is described in Appendix~\ref{app:intervention}.

\subsection{Counterfactuals}
A counterfactual query seeks to quantify statements of the form: what would the value for variable $x_i$ have been if variable $x_j$ had taken value $\alpha$,
\emph{given that we have observed} $\mathbf{x}=\mathbf{x}^{obs}$?  %
The fundamental difference between an interventional and counterfactual query is that the former seeks to marginalize over latent variables, whereas the latter conditions on them.

Given a set of structural equations and an observation $\mathbf{x}^{obs}$, we follow the notation of \citet{pearl2009causality} and write $x_{i,x_j \leftarrow \alpha}(\zb)$
to denote the value of $x_i$ under the counterfactual that $x_j\leftarrow  \alpha$.
As detailed by
\citet{pearl2009causality}, %
counterfactual inference involves three steps: \textit{abduction, action} and \textit{prediction}.
The first step involves,
after fitting the flow to the data,
evaluating the posterior distribution over
latent variables given observations $\mathbf{x}^{obs}$.
This is non-trivial for
most causal models.
However, since flow models readily give access to both forward and backward transformation between observations and latent variables \citep{papamakarios2018masked,kingma2016improving,durkan2019neural}, this first step can be readily evaluated.

The remaining two steps mirror those taken when making
interventional predictions:
the structural equation for the
counterfactual variable is fixed at $\alpha$
and the structural equations are propagated forward.
The only difference here is that the latent samples are drawn from their new distribution:
in fact,
conditioning on $\xb=\xb^{obs}$ changes the distribution of the latent variables by putting a point mass on $\zb = \Tb^{-1}(\xb^{obs})$.
This is summarized in Appendix~\ref{app:counterfactual}.

\section{EXPERIMENTS}

\label{sec:exp}

\subsection{Causal Discovery}

\begin{figure*}
	\begin{center}
		\centerline{\includegraphics[width=2.08\columnwidth]{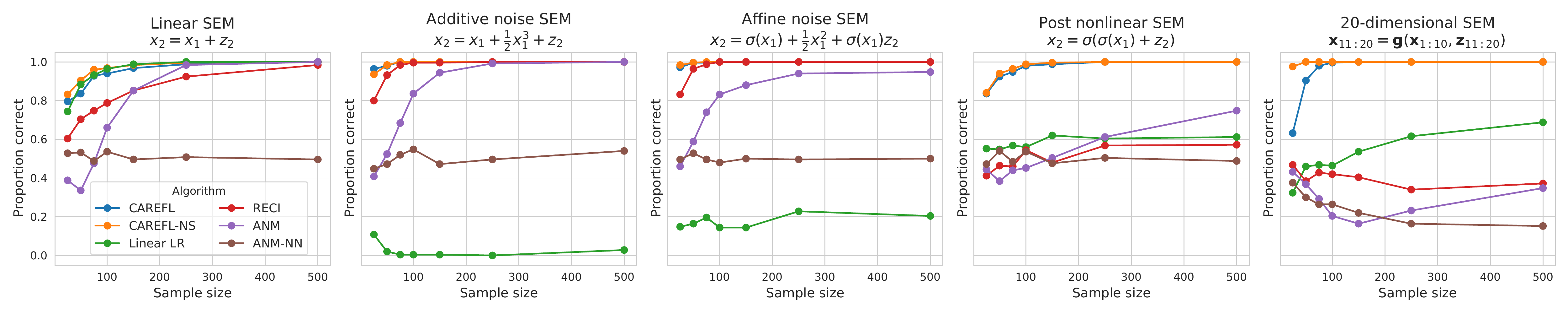}}%
		\caption{Performance on synthetic data generated under distinct SEMs.
		We note that for all five SEMs \carefl performs competitively and is able to robustly identify the underlying causal direction.}
		\label{Fig:causalDiscSimulations}
	\end{center}
\end{figure*}

\begin{figure*}
	\begin{center}
		\centerline{\includegraphics[width=2.08\columnwidth]{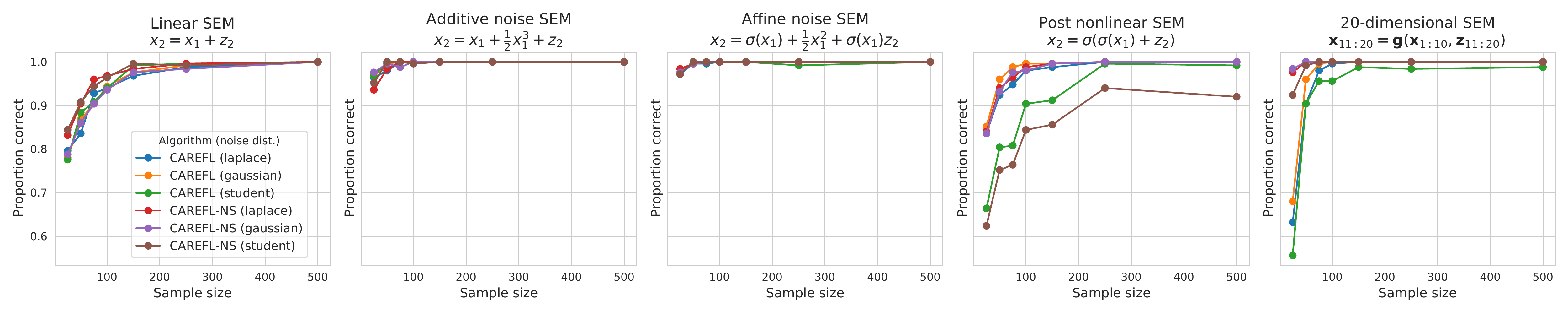}
}%
		\caption{Impact of prior mismatch on the performace of \carefle.
        The prior of each flow is fixed to a Laplace distribution, while the noise distribution is chosen to be either a Laplace, Student-t or Gaussian distribution.}
        \label{fig:prior_mismatch}
	\end{center}
\end{figure*}

We compare the performance of \carefl on a range of synthetic and real world data,
against several alternative methods: the
linear likelihood ratio method of \citet{hyvarinen2013pairwise},
the additive noise model \citep[ANM]{hoyer2009nonlinear,peters2014causal},
and the Regression Error Causal Inference (RECI) method of \citet{bloebaum2018causeeffect}.
For \carefle, we considered the more general affine flows, as well as the special case of additive flows (denoted \carefle-NS, for "non-scaled"), where $s_j = 0$ in~\eqref{eq:flow_def_x_ns}.
For ANM, we considered both a Gaussian process and a neural network as the regression class.
Experimental details can be found in Appendix~\ref{app:exp}.
Code to reproduce the experiments is available \href{https://github.com/piomonti/carefl/}{here}.

\subsubsection{Synthetic data}%

We consider a series of synthetic experiments where the underlying causal model is known.
Data was generated according to the following SEM:
\begin{equation*}
x_1 = z_1 ~~~ \mbox{ and } ~~~ x_2 = f( x_1, z_2)\label{eq:SEMgen}
\end{equation*}
where $z_1, z_2$ follow a standard Laplace distribution.
We consider three distinct forms for $f$:
$(i)$ linear, where $f(x_1, z_2) = \alpha x_1 + z_2$;
$(ii)$ non-linear with additive noise, where $f(x_1, z_2) = x_1 + \alpha x_1^3 + z_2$;
$(iii)$ non-linear with modulated noise, where $f(x_1, z_2) = \sigma(x_1) + \half x_1^2 + \sigma(x_1)z_2$;
$(iv)$ non-linear with non-linear noise, where $f(x_1, z_2) = \sigma \left(\sigma \left( \alpha x_1 \right) + z_2 \right)$.
We write $\sigma$ to denote the sigmoid non-linearity.
We also consider a high dimensional SEM:
\begin{equation*}
\xb_1 = \zb_1 \in \RR^{10} ~~~ \mbox{ and } ~~~ \xb_2 = \gb(\xb_1, \zb_2) \in \RR^{10} \label{eq:SEMgen2}
\end{equation*}
where $\zb_1$ and $\zb_2$ follow standard Laplace distribution, and for each $i\in \iset{1,10}$, $g_i$ has one of the following forms, picked at random:
$(i)$ a function of all inputs $g_i(\xb_1, \zb_2) = \sigma(\sigma(\sum_j x_{1,j}) + z_i)$;
$(ii)$ a function of the first half of the input $g_i(\xb_1, \zb_2) = \sigma(\sigma(\sum_{j\leq5} x_{1,j}) + z_i)$;
$(iii)$ a function of the second half of the input $g_i(\xb_1, \zb_2) = \sigma(\sum_{j > 5} \sigma(x_{1,j})^{j-5} + z_i)$.

For each distinct class of SEMs, we consider the performance of each algorithm under
various distinct sample sizes ranging from $N=25$ to $N=500$ samples.
Furthermore, each experiment is repeated 250 times.
For each repetition, the causal ordering is selected at random.
We implemented \carefl by stacking two affine flows~\eqref{eq:flow_def_x_ns}, where $s_j$ and $t_j$ are feed-forward networks with one hidden layer of dimension 10.

Results are presented in Figure \ref{Fig:causalDiscSimulations}.
Only \carefl is able to consistently uncover the true causal direction in all situations.
We note that the same architecture and training parameters were employed throughout all experiments,
highlighting the fact that the proposed method is agnostic to the nature of the true underlying causal relationship.

We note that while the identifiability results of Theorem \ref{th:iden} are premised on Gaussian noise variables, the simulations used a Laplace distribution instead.
This proves that the Gaussianity assumption is sufficient but not necessary for identifiability to hold.

\paragraph{Robustness to prior misspecification}
In the simulations above, the prior distribution of the flow was chosen to be a Laplace distribution, matching the noise distribution.
To investigate \carefle's robustness to prior mismatch, we run additional simulations where the flow prior is still a Laplace distribution, but the noise distribution is changed.
The remaining of the architectural parameters are kept the same as the simulations above.
The results are shown in Figure~\ref{fig:prior_mismatch}. We see that the performance stays the same.
We also note that in the next subsection, we will consider real world datasets where we did not set the underlying (unknown) noise distribution while maintaining better performance when compared to alternative methods.

\subsubsection{Real data}
\paragraph*{Cause effect pairs data}
We also consider performance of the proposed method on cause-effect pairs benchmark dataset
\citep{mooij2016distinguishing}.
This benchmark consists of  108 distinct
bivariate datasets where the objective is to distinguish between cause and effect.
For each dataset, two separate autoregressive flow models were trained conditional on $\pi=(1,2)$ or $\pi=(2,1)$ and the log-likelihood ratio was evaluated as in equation \eqref{eq:flowLR} to determine the causal variable.
Results are presented in Table~\ref{tab:pairs}.
We note that the proposed method performs better than alternative algorithms.

\begin{table}
	\caption{Percentage of correct
		causal variables identified over 108 pairs from the Cause Effect Pairs benchmark.}
	\label{tab:pairs}
	\vskip 0.15in
	\centering
	\begin{small}
		\begin{sc}
			\begin{tabular}{ccccr}
				\toprule
				CAREFL  & Linear LR & ANM  & RECI \\
				\midrule
				73 $\%$ & 66$\%$ & 69 $\%$   &  69$\%$ \\
				\bottomrule
			\end{tabular}
		\end{sc}
	\end{small}
	\vskip -0.1in
\end{table}

\begin{figure}[!b]
	\begin{center}
		\centerline{\includegraphics[width=1.02\columnwidth]{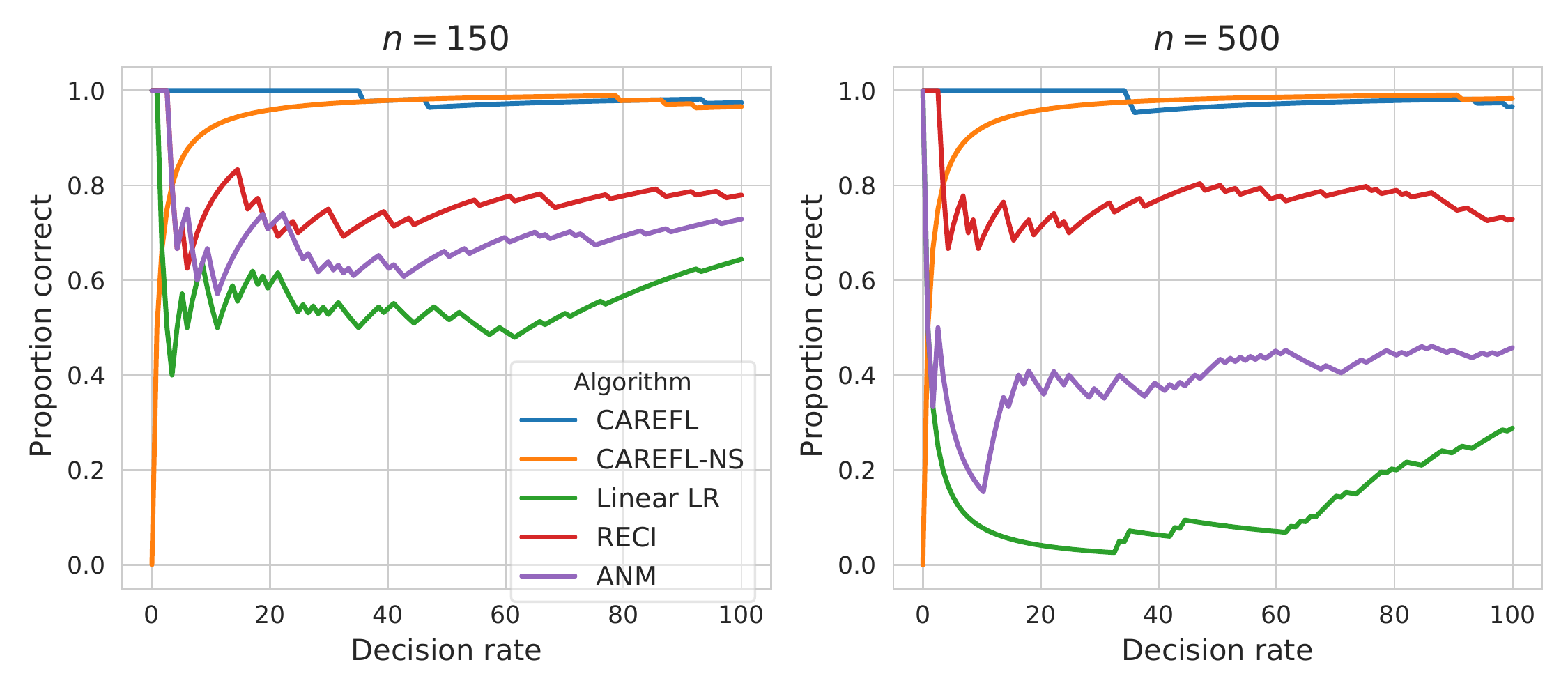}}
		\caption{Performance on finding the arrow of time of EEG data, as a function of decision rate (percentage of channels --- sorted by decreasing confidence --- we have to classify).}
		\label{Fig:eeg}
	\end{center}
\end{figure}

\paragraph*{Arrow of time on EEG data}
Finally, we consider the performance of \carefl in inferring the arrow of time from open-access electroencephalogram (EEG) time series \citep{dornhege2004boosting}.
The data consists of 118 EEG channels for one subject.
We only consider the first $n$ time points, where $n\in\{150, 500\}$,
after which
each of the channels is randomly reversed.
More details on the preprocessing can be found in Appendix~\ref{app:eeg}.
The goal is to correctly infer whether $x_t \rightarrow x_{t+1}$ or $x_{t+1} \rightarrow x_t$ for each channel. This is a useful test case for causal methods since the true direction is known to be from the past to the future.
We report in Figure~\ref{Fig:eeg} the accuracy as a function of the percentage of channels considered, sorted from highest to lowest confidence (\ie by how high the amplitude of the output of each algorithm is).
For average to high confidence, \carefl is comparable in performance to the baseline methods, but performs better in the low confidence regime.
We also note that the performance of \carefl improves by increasing the sample-size, which is to be expected from a method based on deep learning.

\subsection{Interventions}
To demonstrate that \carefl can answer interventional queries, we will consider both a synthetic controlled example, as well as real fMRI data.

\paragraph*{Synthetic data}
Consider four-dimensional data generated as
\begin{align}
\begin{split}
\label{intervention_SEM}
x_1& = z_1 \qquad x_3 = x_1 + c_1 x_2^3 + z_3\\
x_2& = z_2 \qquad x_4 = c_2 x_1^2 - x_2 + z_4
\end{split}
\end{align}
where each $z_i$ is drawn independently from a standard Laplace distribution, and $(c_1, c_2)$ are random coefficients.
From the SEM above we can derive the expectations for $x_3$ and $x_4$ under
an intervention  $do(X_1=\alpha)$ as being $\alpha$ and $c_2 \alpha^2 $ respectively.

\begin{figure}[t!]
	\begin{center}
		\centerline{\includegraphics[width=1.02\columnwidth]{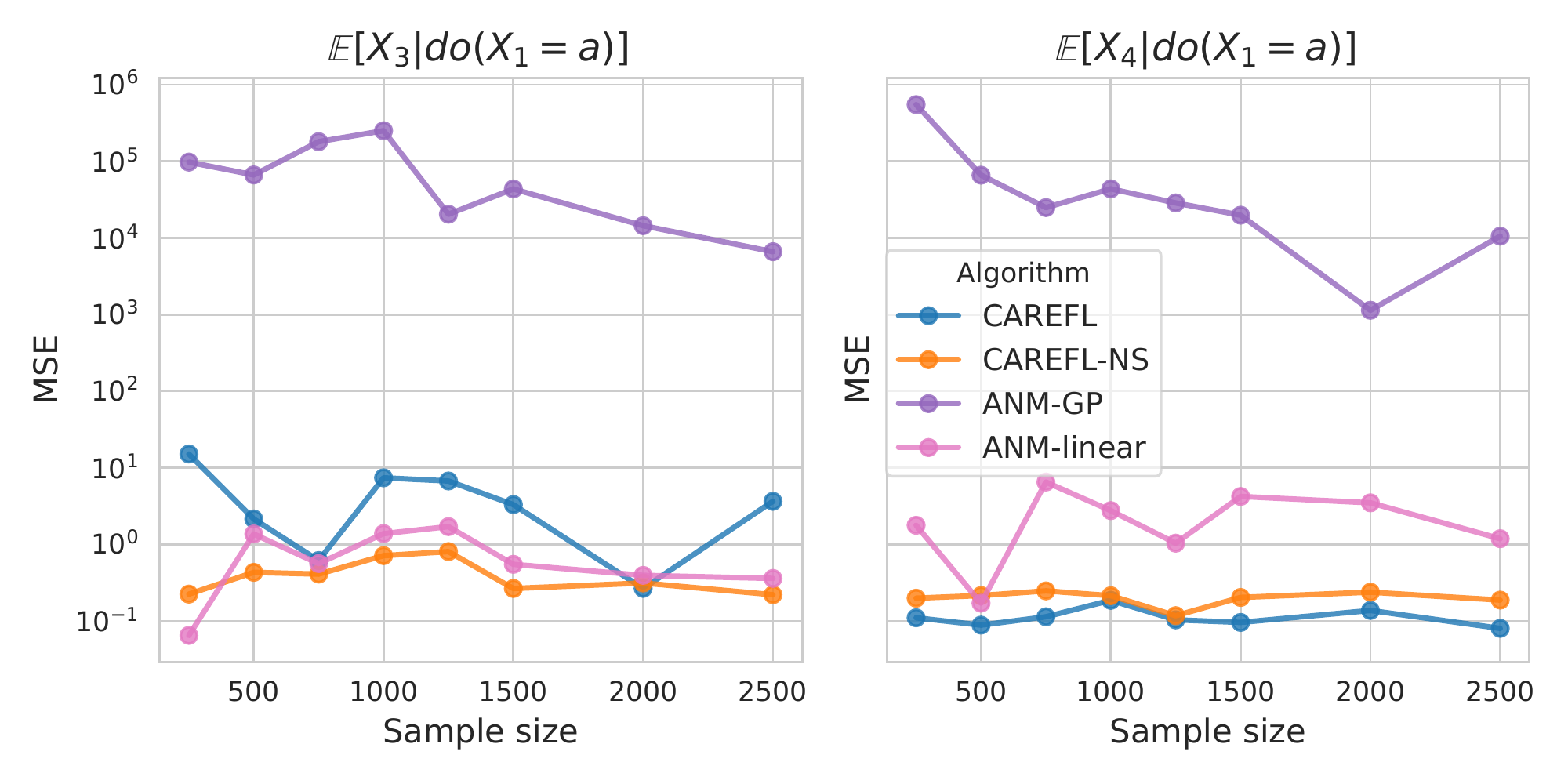}}
		\caption{Mean square error for interventional predictions on simulated data, generated using equation (\ref{intervention_SEM}).
        The left and right panels consider
        linear and non-linear interventional distributions.
		}
		\label{Fig:interventionExample1}
	\end{center}
	\vskip -0.2in
\end{figure}

We compare \carefl against the regression function from an ANM \citep{hoyer2009nonlinear}, where the regression is either linear or a Gaussian process.
Figure \ref{Fig:interventionExample1} visualizes the expected mean squared error between
predicted expectations for $x_3$ and $x_4$ under
the intervention $do( X_1=\alpha)$ for the proposed method,
and the true expectations.
We note that
\carefl is able to better infer the nature of
the true interventional distributions when compared to the baseline.

\paragraph*{Interventional fMRI data}
In order to validate the performance on  interventional real-data we applied \carefl to open-access electrical stimulation fMRI \citep{thompson2020human}.
Data was collected across 26 patients with
medically refractory epilepsy,
which required
surgically
implanting intracranial electrodes in cortical and subcortical locations. FMRI data was then collected
during rest as well as while electrodes were being stimulated.
Whilst each patient had electrodes implanted in slightly different locations, we identified 16 patients with electrodes in or near the Cingulate Gyrus
and studied these patients exclusively. We further restricted ourselves
to studying the data from the Cingulate Gyrus (CG)
and
Heschl's Gyrus (HG), resulting in bivariate time-series per patient.
Full data preprocessing and preparation is described in Appendix~\ref{app:esfmri}.

We compared \carefl with both linear and additive noise models.
Throughout these experiments we
assumed the underlying causal
structure between regions was known (with CG$\rightarrow$HG) and trained each
model using the resting-state data. Given the trained model,
sessions where the CG was stimulated were treated as
interventional sessions, with the task being to
predict fMRI activation in HG given CG activity.
Whilst the true underlying DAG will be certainly be more complex than
the simple bivariate structure considered here, these experiments
nonetheless serve as a real dataset benchmark through which to compare
various causal inference algorithms. The results are provided in
Table~\ref{tab:esfmri}, where \carefl is shown to
out-perform alternative causal models.

\begin{table}[!ht]
	\caption{Median absolute error for interventional predictions in electrical stimulation fMRI data.}
	\label{tab:esfmri}
	\centering
	\begin{small}
		\begin{sc}
		\begin{tabular}{lc}
			\toprule
			Algorithm  & Median abs error (std. dev.) \\
			\midrule
		    \carefle   &  \textbf{0.586 (0.048)} \\
		    ANM  & 0.655 (0.057) \\
		    Linear SEM & 0.643 (0.044) \\
			\bottomrule
		\end{tabular}
		\end{sc}
	\end{small}
\end{table}

\subsection{Counterfactuals}
\begin{figure}[!ht]
    \vskip 0.2in
    \begin{center}
        \centerline{\includegraphics[width=\columnwidth]{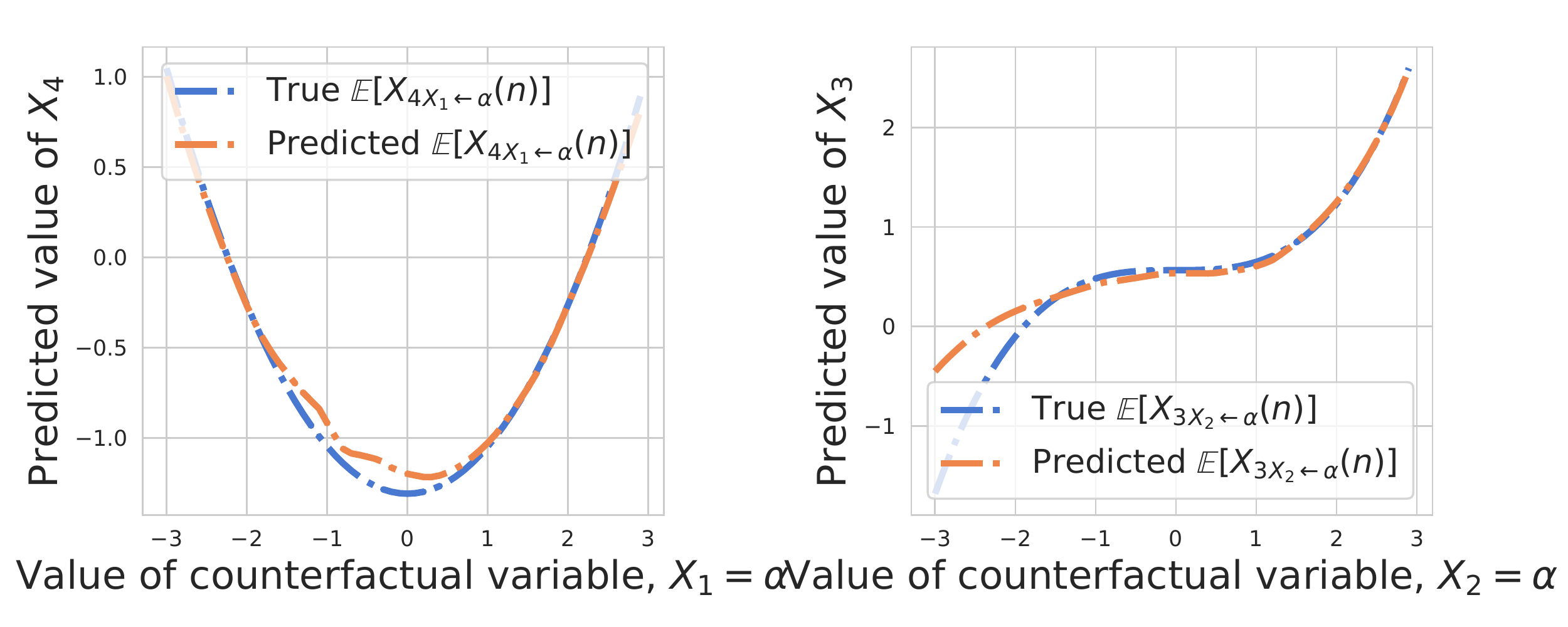}}
        \caption{Counterfactual predictions for variables $x_3$ and $x_4$. Note that
            flow is able to obtain accurate counterfactual predictions for a
            range of values of $\alpha$.} %
        \label{Fig:counterfactualExample1}
    \end{center}
    \vskip -0.2in
\end{figure}

We continue with the simple 4 dimensional structural equation model
described in equation (\ref{intervention_SEM}).
We assume we observe $\mathbf{x}^{obs} = (2.00  ,  1.50 ,  0.81, -0.28)$
and consider the counterfactual
values under two distinct scenarios:
$(i)$ the expected counterfactual value of $x_3$ if $x_2=\alpha$ instead of $x_1=2$;
$(ii)$ the expected counterfactual value of $x_4$ if $x_1=\alpha$ instead of $x_1=2$.
Counterfactual predictions
require us to infer the values of latent
variables, called \textit{abduction} step by
\cite{pearl2009causal}. This is non-trivial for
most causal models, but can be easily
achieved with \carefl due
to the invertibility of flow models.
Figure~\ref{Fig:counterfactualExample1}
demonstrates that
\carefl can indeed make accurate counterfactual predictions. %

\section{DISCUSSION}

Existing identifiability results on causal models other than additive noise models are limited.
To our knowledge,
the other notable and identifiable non-additive noise models are the post-non-linear model \citep[PNL]{zhang2009identifiability} and the non-stationary non-linear SEM model \citep[NonSENS]{monti2019causal}.
The PNL model assumes that the cause $x$ and the effect $y$ are related through the equation $y = f_2(f_1(x) + n)$, where $n$ is a noise variable independent of $x$.
In contrast to affine flows, the function $f_2$ is fixed (in the sense of not modulated by the cause $x$), while being non-linear as opposed to affine.
By applying its inverse $f_2^{-1}$ to $y$, we actually end up with an additive noise model.

In our model, in stark contrast to the PNL model, it is not possible to apply a fixed (as in not a function of the cause) transformation to the effect to revert back to an additive noise model.
This is the main reason why the existing identifiability theory doesn't cover our causal model~\eqref{eq:flow_def_x_ns}.
Theorem~\ref{th:iden} thus presents a novel identifiability result in the context of non-additive noise models, and the proposed estimation algorithm benefits from it, as was shown in our experiments.

The NonSENS framework allows for general non-linear relationships between cause, noise and effect. Assuming access to non-stationary data, it is identifiable even in such a general case by leveraging recent
results in the theory of non-linear ICA  \citep{hyvarinen2016unsupervised,hyvarinen2019nonlinear,khemakhem2020icebeem,khemakhem2020variational}.
In contrast, the proposed model does restrict the nature of non-linear relationships but places no assumptions of nonstationarity, so our model can be applied in more general scenarios.
Our work follows a recent trend of combining flexible generative models (such as autoregressive flows and VAEs) with structural causal models \citep{pawlowski2020deep,wehenkel2020graphical,louizos2017causal}.

In the context of additive noise models, the estimation methods by \citet[ANM]{hoyer2009nonlinear} and \citet[RECI]{bloebaum2018causeeffect} require least-squares regressions in both directions.
RECI then compares the magnitudes of the residuals, while ANM depends on independence tests between residuals and causes.
Choosing the right regression model in both these methods is difficult. As stated by \cite{bloebaum2018causeeffect}, a very good regression function can reduce the performance of ANM and RECI because it decreases the confidence of the independence tests.
We have observed this in our experiments when using neural networks as the regression class, as seen in Figure~\Ref{Fig:causalDiscSimulations}.
Importantly, if the additive noise assumption fails to hold, both approaches will fail regardless of the regression class.

\carefl is specifically leveraging the recent developments in deep learning with the promise of finding computationally efficient methods, as well as improving the statistical efficiency (power) by using likelihood ratios.
Furthermore, both ANM and RECI were solely designed for causal discovery,
and the invertibility of the system in order to perform interventions and counterfactuals wasn't discussed. So, it is plausible that our model might be preferable even the context of ANM's, in addition to generalizing them.

We note that the likelihood ratio approach by  \citet{hyvarinen2013pairwise} was originally designed for LiNGAM, which is a linear model based on non-Gaussianity \citep{shimizu2006linear}. An extension of likelihood ratios to non-linear ANM was also proposed by \citet{hyvarinen2013pairwise}, together with a heuristic approximation which roughly amounts to RECI.

\section{CONCLUSION}

We argue that autoregressive flow models are well-suited to causal inference tasks, ranging from causal discovery to making counterfactual predictions. This is because we can interpret the ordering of variables in an autoregressive flow in the framework of SEMs.

We show that affine flows in particular define a new class of causal models, where the noise is modulated by the cause. For such models, we prove a completely new causal identifiability result which generalizes additive noise models.
We show how to efficiently learn causal structure by selecting the ordering with the highest test log-likelihood
and thus present a measure of causal direction based on the likelihood-ratio for non-linear SEMs.

Furthermore, by restricting ourselves to autoregressive flow models we are able to easily evaluate interventional queries by fixing
the interventional variable whilst sampling from the flow. The invertible property of autoregressive flows further facilitates the evaluation of counterfactual queries.

In experiments on synthetic and real data, our method outperformed
alternative methods in causal discovery as well as interventional and counterfactual predictions.

\ackaccepted

I.K. and R.P.M.  were supported by the Gatsby Charitable Foundation. A.H. was supported by a Fellowship from CIFAR.
\newpage

\bibliography{refs}

\begin{thebibliography}{}

\bibitem[Abraham et~al., 2014]{abraham2014machine}
Abraham, A., Pedregosa, F., Eickenberg, M., Gervais, P., Mueller, A., Kossaifi,
  J., Gramfort, A., Thirion, B., and Varoquaux, G. (2014).
\newblock Machine learning for neuroimaging with scikit-learn.
\newblock {\em Frontiers in neuroinformatics}, 8:14.

\bibitem[Bloebaum et~al., 2018]{bloebaum2018causeeffect}
Bloebaum, P., Janzing, D., Washio, T., Shimizu, S., and Sch{\"o}lkopf, B.
  (2018).
\newblock Cause-effect inference by comparing regression errors.
\newblock In {\em International {{Conference}} on {{Artificial Intelligence}}
  and {{Statistics}}}, pages 900--909.

\bibitem[Dinh et~al., 2014]{dinh2014nice}
Dinh, L., Krueger, D., and Bengio, Y. (2014).
\newblock {{NICE}}: {{Non}}-linear {{Independent Components Estimation}}.
\newblock {\em arXiv:1410.8516 [cs]}.

\bibitem[Dinh et~al., 2016]{dinh2016density}
Dinh, L., {Sohl-Dickstein}, J., and Bengio, S. (2016).
\newblock Density estimation using {{Real NVP}}.
\newblock {\em arXiv:1605.08803 [cs, stat]}.

\bibitem[Dornhege et~al., 2004]{dornhege2004boosting}
Dornhege, G., Blankertz, B., Curio, G., and M{\"u}ller, K.-R. (2004).
\newblock Boosting bit rates in noninvasive {{EEG}} single-trial
  classifications by feature combination and multiclass paradigms.
\newblock {\em IEEE transactions on bio-medical engineering}, 51(6):993--1002.

\bibitem[Durkan et~al., 2019a]{durkan2019cubicspline}
Durkan, C., Bekasov, A., Murray, I., and Papamakarios, G. (2019a).
\newblock Cubic-{{Spline Flows}}.
\newblock {\em arXiv:1906.02145 [cs, stat]}.

\bibitem[Durkan et~al., 2019b]{durkan2019neural}
Durkan, C., Bekasov, A., Murray, I., and Papamakarios, G. (2019b).
\newblock Neural {{Spline Flows}}.
\newblock {\em arXiv:1906.04032 [cs, stat]}.

\bibitem[Esteban et~al., 2019]{esteban2019fmriprep}
Esteban, O., Markiewicz, C.~J., Blair, R.~W., Moodie, C.~A., Isik, A.~I.,
  Erramuzpe, A., Kent, J.~D., Goncalves, M., DuPre, E., Snyder, M., et~al.
  (2019).
\newblock fmriprep: a robust preprocessing pipeline for functional mri.
\newblock {\em Nature methods}, 16(1):111--116.

\bibitem[Germain et~al., 2015]{germain2015made}
Germain, M., Gregor, K., Murray, I., and Larochelle, H. (2015).
\newblock {{MADE}}: {{Masked Autoencoder}} for {{Distribution Estimation}}.
\newblock {\em arXiv:1502.03509 [cs, stat]}.

\bibitem[Gorgolewski et~al., 2011]{gorgolewski2011nipype}
Gorgolewski, K., Burns, C.~D., Madison, C., Clark, D., Halchenko, Y.~O.,
  Waskom, M.~L., and Ghosh, S.~S. (2011).
\newblock Nipype: a flexible, lightweight and extensible neuroimaging data
  processing framework in python.
\newblock {\em Frontiers in neuroinformatics}, 5:13.

\bibitem[Hornik, 1991]{hornik1991approximation}
Hornik, K. (1991).
\newblock Approximation capabilities of multilayer feedforward networks.
\newblock {\em Neural Networks}, 4(2):251--257.

\bibitem[Hoyer et~al., 2009]{hoyer2009nonlinear}
Hoyer, P.~O., Janzing, D., Mooij, J.~M., Peters, J., and Sch{\"o}lkopf, B.
  (2009).
\newblock Nonlinear causal discovery with additive noise models.
\newblock In {\em Advances in Neural Information Processing Systems}, pages
  689--696.

\bibitem[Huang et~al., 2018]{huang2018neural}
Huang, C.-W., Krueger, D., Lacoste, A., and Courville, A. (2018).
\newblock Neural {{Autoregressive Flows}}.
\newblock {\em arXiv:1804.00779 [cs, stat]}.

\bibitem[Hyv{\"a}rinen and Morioka, 2016]{hyvarinen2016unsupervised}
Hyv{\"a}rinen, A. and Morioka, H. (2016).
\newblock Unsupervised feature extraction by time-contrastive learning and
  nonlinear {{ICA}}.
\newblock In {\em Advances in {{Neural Information Processing Systems}}}, pages
  3765--3773.

\bibitem[Hyv{\"a}rinen and Pajunen, 1999]{hyvarinen1999nonlinear}
Hyv{\"a}rinen, A. and Pajunen, P. (1999).
\newblock Nonlinear independent component analysis: {{Existence}} and
  uniqueness results.
\newblock {\em Neural Networks}, 12(3):429--439.

\bibitem[Hyv{\"a}rinen et~al., 2019]{hyvarinen2019nonlinear}
Hyv{\"a}rinen, A., Sasaki, H., and Turner, R. (2019).
\newblock Nonlinear {{ICA Using Auxiliary Variables}} and {{Generalized
  Contrastive Learning}}.
\newblock In {\em The 22nd {{International Conference}} on {{Artificial
  Intelligence}} and {{Statistics}}}, pages 859--868.

\bibitem[Hyv{\"a}rinen and Smith, 2013]{hyvarinen2013pairwise}
Hyv{\"a}rinen, A. and Smith, S.~M. (2013).
\newblock Pairwise {{Likelihood Ratios}} for {{Estimation}} of
  {{Non}}-{{Gaussian Structural Equation Models}}.
\newblock {\em Journal of Machine Learning Research}, 14(Jan):111--152.

\bibitem[Khemakhem et~al., 2020a]{khemakhem2020variational}
Khemakhem, I., Kingma, D.~P., Monti, R.~P., and Hyv{\"a}rinen, A. (2020a).
\newblock Variational {{Autoencoders}} and {{Nonlinear ICA}}: {{A Unifying
  Framework}}.
\newblock In {\em The 23rd {{International Conference}} on {{Artificial
  Intelligence}} and {{Statistics}}}.

\bibitem[Khemakhem et~al., 2020b]{khemakhem2020icebeem}
Khemakhem, I., Monti, R.~P., Kingma, D.~P., and Hyv{\"a}rinen, A. (2020b).
\newblock {{ICE}}-{{BeeM}}: {{Identifiable Conditional Energy}}-{{Based Deep
  Models}}.
\newblock {\em arXiv:2002.11537 [cs, stat]}.

\bibitem[Kingma and Dhariwal, 2018]{kingma2018glow}
Kingma, D.~P. and Dhariwal, P. (2018).
\newblock Glow: {{Generative Flow}} with {{Invertible}} 1x1 {{Convolutions}}.
\newblock {\em arXiv:1807.03039 [cs, stat]}.

\bibitem[Kingma et~al., 2016]{kingma2016improving}
Kingma, D.~P., Salimans, T., Jozefowicz, R., Chen, X., Sutskever, I., and
  Welling, M. (2016).
\newblock Improving {{Variational Inference}} with {{Inverse Autoregressive
  Flow}}.
\newblock {\em arXiv:1606.04934 [cs, stat]}.

\bibitem[Kobyzev et~al., 2020]{kobyzev2020normalizing}
Kobyzev, I., Prince, S. J.~D., and Brubaker, M.~A. (2020).
\newblock Normalizing {{Flows}}: {{An Introduction}} and {{Review}} of
  {{Current Methods}}.
\newblock {\em IEEE Transactions on Pattern Analysis and Machine Intelligence},
  pages 1--1.

\bibitem[Louizos et~al., 2017]{louizos2017causal}
Louizos, C., Shalit, U., Mooij, J., Sontag, D., Zemel, R., and Welling, M.
  (2017).
\newblock Causal effect inference with deep latent-variable models.
\newblock {\em arXiv preprint arXiv:1705.08821}.

\bibitem[Monti et~al., 2019]{monti2019causal}
Monti, R.~P., Zhang, K., and Hyvarinen, A. (2019).
\newblock Causal discovery with general non-linear relationships using
  non-linear {{ICA}}.
\newblock In {\em 35th {{Conference}} on {{Uncertainty}} in {{Artificial
  Intelligence}}, {{UAI}} 2019}, volume~35.

\bibitem[Mooij et~al., 2016]{mooij2016distinguishing}
Mooij, J.~M., Peters, J., Janzing, D., Zscheischler, J., and Sch{\"o}lkopf, B.
  (2016).
\newblock Distinguishing cause from effect using observational data: Methods
  and benchmarks.
\newblock {\em The Journal of Machine Learning Research}, 17(1):1103--1204.

\bibitem[Neyman and Pearson, 1933]{neyman1933ix}
Neyman, J. and Pearson, E.~S. (1933).
\newblock {{IX}}. {{On}} the problem of the most efficient tests of statistical
  hypotheses.
\newblock {\em Philosophical Transactions of the Royal Society of London.
  Series A, Containing Papers of a Mathematical or Physical Character},
  231(694-706):289--337.

\bibitem[Papamakarios et~al., 2019]{papamakarios2019normalizing}
Papamakarios, G., Nalisnick, E., Rezende, D.~J., Mohamed, S., and
  Lakshminarayanan, B. (2019).
\newblock Normalizing flows for probabilistic modeling and inference.
\newblock {\em arXiv preprint arXiv:1912.02762}.

\bibitem[Papamakarios et~al., 2018]{papamakarios2018masked}
Papamakarios, G., Pavlakou, T., and Murray, I. (2018).
\newblock Masked {{Autoregressive Flow}} for {{Density Estimation}}.
\newblock {\em arXiv:1705.07057 [cs, stat]}.

\bibitem[Pawlowski et~al., 2020]{pawlowski2020deep}
Pawlowski, N., Castro, D.~C., and Glocker, B. (2020).
\newblock Deep structural causal models for tractable counterfactual inference.
\newblock {\em arXiv preprint arXiv:2006.06485}.

\bibitem[Pearl, 2009a]{pearl2009causal}
Pearl, J. (2009a).
\newblock Causal inference in statistics: {{An}} overview.
\newblock {\em Statistics Surveys}, 3:96--146.

\bibitem[Pearl, 2009b]{pearl2009causality}
Pearl, J. (2009b).
\newblock {\em Causality}.
\newblock {Cambridge University Press}, {Cambridge}.

\bibitem[Peters et~al., 2016]{peters2016causal}
Peters, J., B{\"u}hlmann, P., and Meinshausen, N. (2016).
\newblock Causal inference by using invariant prediction: Identification and
  confidence intervals.
\newblock {\em Journal of the Royal Statistical Society: Series B (Statistical
  Methodology)}, 78(5):947--1012.

\bibitem[Peters et~al., 2014]{peters2014causal}
Peters, J., Mooij, J., Janzing, D., and Sch{\"o}lkopf, B. (2014).
\newblock Causal {{Discovery}} with {{Continuous Additive Noise Models}}.
\newblock {\em arXiv:1309.6779 [stat]}.

\bibitem[Rezende and Mohamed, 2015]{rezende2015variational}
Rezende, D.~J. and Mohamed, S. (2015).
\newblock Variational {{Inference}} with {{Normalizing Flows}}.
\newblock {\em arXiv:1505.05770 [cs, stat]}.

\bibitem[Shimizu et~al., 2006]{shimizu2006linear}
Shimizu, S., Hoyer, P.~O., Hyv{\"a}rinen, A., and Kerminen, A. (2006).
\newblock A {{Linear Non}}-{{Gaussian Acyclic Model}} for {{Causal Discovery}}.
\newblock {\em Journal of Machine Learning Research}, 7(Oct):2003--2030.

\bibitem[Shimizu et~al., 2011]{shimizu2011directlingam}
Shimizu, S., Inazumi, T., Sogawa, Y., Hyv{\"a}rinen, A., Kawahara, Y., Washio,
  T., Hoyer, P.~O., and Bollen, K. (2011).
\newblock {{DirectLiNGAM}}: {{A}} direct method for learning a linear
  non-{{Gaussian}} structural equation model.
\newblock {\em Journal of Machine Learning Research}, 12(Apr):1225--1248.

\bibitem[Spirtes et~al., 2000]{spirtes2000causation}
Spirtes, P., Glymour, C.~N., Scheines, R., and Heckerman, D. (2000).
\newblock {\em Causation, Prediction, and Search}.
\newblock {MIT press}.

\bibitem[Spirtes and Zhang, 2016]{spirtes2016causal}
Spirtes, P. and Zhang, K. (2016).
\newblock Causal discovery and inference: Concepts and recent methodological
  advances.
\newblock {\em Applied Informatics}, 3(1):3.

\bibitem[Thompson et~al., 2020]{thompson2020human}
Thompson, W.~H., Nair, R., Oya, H., Esteban, O., Shine, J.~M., Petkov, C.,
  Poldrack, R.~A., Howard, M., and Adolphs, R. (2020).
\newblock Human es-{{fMRI Resource}}: {{Concurrent}} deep-brain stimulation and
  whole-brain functional {{MRI}}.
\newblock {\em bioRxiv}, page 2020.05.18.102657.

\bibitem[Vogt, 2019]{vogt2019cingulate}
Vogt, B.~A. (2019).
\newblock {\em Cingulate Cortex}.
\newblock Elsevier.

\bibitem[Wehenkel and Louppe, 2020]{wehenkel2020graphical}
Wehenkel, A. and Louppe, G. (2020).
\newblock Graphical normalizing flows.
\newblock {\em arXiv preprint arXiv:2006.02548}.

\bibitem[Zhang et~al., 2017]{zhang2017causal}
Zhang, K., Huang, B., Zhang, J., Glymour, C., and Sch{\"o}lkopf, B. (2017).
\newblock Causal discovery from nonstationary/heterogeneous data: {{Skeleton}}
  estimation and orientation determination.
\newblock In {\em {{IJCAI}}: {{Proceedings}} of the {{Conference}}}, volume
  2017, page 1347. {NIH Public Access}.

\bibitem[Zhang and Hyvarinen, 2009]{zhang2009identifiability}
Zhang, K. and Hyvarinen, A. (2009).
\newblock On the identifiability of the post-nonlinear causal model.
\newblock In {\em 25th {{Conference}} on {{Uncertainty}} in {{Artificial
  Intelligence}}, {{UAI}} 2009}, volume~35.

\bibitem[Zhang et~al., 2015a]{zhang2015estimation}
Zhang, K., Wang, Z., Zhang, J., and Sch{\"o}lkopf, B. (2015a).
\newblock On {{Estimation}} of {{Functional Causal Models}}: {{General
  Results}} and {{Application}} to the {{Post}}-{{Nonlinear Causal Model}}.
\newblock {\em ACM Transactions on Intelligent Systems and Technology},
  7(2):13:1--13:22.

\bibitem[Zhang et~al., 2015b]{zhang2015distinguishing}
Zhang, K., Zhang, J., and Sch{\"o}lkopf, B. (2015b).
\newblock Distinguishing {{Cause}} from {{Effect Based}} on {{Exogeneity}}.
\newblock {\em arXiv:1504.05651 [cs, stat]}.

\bibitem[Zheng et~al., 2018]{zheng2018dags}
Zheng, X., Aragam, B., Ravikumar, P.~K., and Xing, E.~P. (2018).
\newblock {{DAGs}} with {{NO TEARS}}: {{Continuous}} optimization for structure
  learning.
\newblock In {\em Advances in {{Neural Information Processing Systems}}}, pages
  9472--9483.

\end{thebibliography}

\onecolumn
\newpage
\appendix

\begin{center}
    \Large \bf Appendix
\end{center}

\section{Identifiability of the affine causal model}
\label{app:iden}
Recall the form of the SEM that is defined by an autoregressive affine flow:
\begin{equation}
    \label{eq:flow_def_app}
    x_j = e^{s_j(x_{<\pi(j)})}z_j + t_j(x_{<\pi(j)}),\quad j=1,2
\end{equation}
where $\pi$ is a permutation that describes the causal ordering.

The proof for additive flows ($s_1 = s_2 = 0$ in equation~\eqref{eq:flow_def_app}) and general noise can be found in~\cite{hoyer2009nonlinear}.
Theorem~\ref{th:iden_gen} below summarizes the two scenarios under which the causal model defined by an affine flow is not identifiable. In particular, if the function $t_j$ in equation~\eqref{eq:flow_def_app} linking cause to effect is invertible and non-linear, then none of these scenarios can hold.
In addition, the proof of Theorem~\ref{th:iden_gen} only requires one of the noise variables to be Gaussian.

\begin{definition}
\label{def:lograt}
Let $(\alpha, \gamma, \delta, \beta, \alpha_0, \beta_0, \gamma_0, \delta_0) \in \RR_{\geq 0}\times\RR_{>0}^2\times\RR^5$ be a tuple such that one of the following conditions holds:
\begin{itemize}
\item $\alpha > 0$,  $\alpha_0^2  < \alpha \delta$ and  $\beta^2 < 4\alpha\gamma$.
\item $\alpha=\beta=\alpha_0 = 0$ and $\beta_0^2 < \delta $.
\end{itemize}
We say that a density $p_x$ of a continuous variable $x$ is \emph{log-mix-rational-log} if it has the form:
\begin{equation}
    \log p_x(x) = - \half \delta x^2 + \delta_0x + \half \frac{\left(\alpha_0 x^2 + \beta_0 x + \gamma_0\right)^2}{\alpha x^2 + \beta x + \gamma} - \half \log(\alpha x^2 + \beta x + \gamma)  + \textrm{const}
\end{equation}
We say that $p_x$ is \emph{strictly} log-mix-rational-log if $\alpha > 0$.
\end{definition}

Note that the Gaussian distribution is part of the log-mix-rational-log family, for $\alpha = \beta = \alpha_0 = 0$. If $\alpha \neq 0$, then the log-mix-rational-log family is not part of the exponential family.

\begin{theorem}
\label{th:iden_gen}
Assume the data follows the model
\begin{equation} \label{modmodelxy}
y=f(x)+v(x) n
\end{equation}
where $n$ is a standardized Gaussian independent of $x$, $f$ and $v$ are twice-differentiable scalar functions defined on $\RR$ and $v > 0$.\\
If a backward model exists, \ie the data also follows the same model in the other direction
\begin{equation} \label{modmodelyx}
x=g(y)+w(y) m
\end{equation}
where $m$ is a standardized Gaussian independent of $y$ and $w>0$, then one of the following scenarios must hold:
\begin{enumerate}
    \item \label{scenario1} $(v, f) = \left(\frac{1}{Q}, \frac{P}{Q}\right)$ and $(w, g) = \left(\frac{1}{Q'}, \frac{P'}{Q'}\right)$ where $Q, Q'$ are polynomials of degree two, $Q, Q' > 0$, $P, P'$ are polynomials of degree two or less, and $p_x, p_y$ are strictly log-mix-rational-log.
    In particular, $\lim_{-\infty} v = \lim_{+\infty}v = 0^+$, $\lim_{-\infty} f = \lim_{+\infty}f < \infty$, $\lim_{-\infty} w = \lim_{+\infty}w = 0^+$, $\lim_{-\infty} g = \lim_{+\infty}g < \infty$ and $f, v, g, w$ are not invertible.
    \item \label{scenario2} $v, w$ are constant, $f, g$ are linear and $p_x, p_y$ are Gaussian densities.
\end{enumerate}
\end{theorem}

\newcommand{\vinv}{\overline{v}}
\newcommand{\winv}{\overline{w}}

\textit{Proof.}
The log-likelihood of \eqref{modmodelxy}, denoted by $p_1$, is given by
\begin{equation}
    \log p_1(x,y)=\log p_x(x) -\half \left(\frac{y-f(x)}{v(x)}\right)^2 -  \log v(x)-\half \log 2\pi
\end{equation}
and log-likelihood of \eqref{modmodelyx}, denoted by $p_2$, is given by
\begin{equation}
    \log p_2(x,y)=\log p_y(y) -\half \left(\frac{x-g(y)}{w(y)}\right)^2 - \log w(y) -\half \log 2\pi
\end{equation}
If the data follows both models, these are equal:
\begin{equation} \label{modmodelequality}
    \log p_x(x) -\half \left(\frac{y-f(x)}{v(x)}\right)^2 -  \log v(x)
    =
    \log p_y(y) -\half \left(\frac{x-g(y)}{w(y)}\right)^2 - \log w(y)
\end{equation}
Denote $\frac{1}{v(x)}$ by $\vinv(x)$ and likewise for $w$.
Now, take the derivative of both sides with respect to $x$:
\begin{equation}
    (\log p_x)'(x) -  \vinv(x)(y-f(x)) (y\vinv'(x)-(f\vinv)'(x))  - (\log v)'(x)
    = - (x-g(y)) \winv^2(y)
\end{equation}
Take the derivative of both sides of this with respect to $y$:
\begin{equation}
\label{modmodeltemp1}
    -  \vinv(x)[2y\vinv'(x)-(f\vinv)'(x)- f(x)\vinv'(x)]
    = - x (\winv^2)'(y)  +g'(y) \winv^2(y) +g(y) (\winv^2)'(y)
\end{equation}
Again, take the derivative of both sides with respect to $x$:
\begin{equation} \label{modmodeltemp}
    -  y(\vinv^2)''(x)+[\vinv((f\vinv)'+ f\vinv')]'(x) = -  (\winv^2)'(y)
\end{equation}
and once more,  take the derivative of both sides of this with respect to $y$:
\begin{equation}
    -  (\vinv^2)''(x)  = -  (\winv^2)''(y)
\end{equation}
which is possible only if both sides are constant, which is equivalent to $\vinv^2$ and $\winv^2$ being second-order polynomials.
In other words,
\begin{equation}
\label{eq:vinv}
    \vinv^2(x)=\alpha x^2 + \beta x + \gamma, \:\:\: v^2(x)=\frac{1}{\alpha x^2 + \beta x + \gamma}
\end{equation}
where the parameters must be such that the $\vinv$ is always positive. The same holds for $w$:
\begin{equation}
\label{eq:winv}
    \winv^2(y)=\alpha' y^2 + \beta' y + \gamma', \:\:\: w^2(y)=\frac{1}{\alpha' y^2 + \beta' y + \gamma'}
\end{equation}

Furthermore, equation~\eqref{modmodeltemp} together with the fact that $(\vinv^2)''(x) = \textrm{const}$ implies that:
\begin{equation}
    [\vinv((f\vinv)'+f\vinv')]'(x) =[f'\vinv^2+2f(\vinv^2)')]'(x) =(f\vinv^2)''(x)= \text{const}
\end{equation}
or
\begin{equation}
    f(x)\vinv^2(x)=\alpha_0 x^2 + \beta_0 x + \gamma_0
\end{equation}
which means that $f$ has the following form:
\begin{equation}
\label{eq:f}
    f(x)=\frac{\alpha_0 x^2 + \beta_0 x + \gamma_0}{\alpha x^2 + \beta x + \gamma}
\end{equation}
The same analysis yields a similar form for $g$:
\begin{equation}
\label{eq:g}
    g(y)=\frac{\alpha_0' y^2 + \beta_0' y + \gamma_0'}{\alpha' x^2 + \beta' x + \gamma'}
\end{equation}

For $\vinv$ to be always positive, the coefficients $(\alpha, \beta, \gamma)$ in~\eqref{eq:vinv} must satisfy one of the following conditions:
\begin{enumerate}[label=\arabic*.]
    \item \label{c1} $\alpha > 0$ and $4\alpha\gamma - \beta^2 > 0$.
    \item \label{c2} $\alpha = \beta = 0$ and $\gamma > 0$.
\end{enumerate}

Similarly, for $\winv$ to be always positive, the coefficients $(\alpha', \beta', \gamma')$ in~\eqref{eq:winv} must satisfy one of the following conditions:
\begin{enumerate}[label=\arabic*'.]
    \item \label{c1p} $\alpha' > 0$ and $4\alpha'\gamma' - \beta'^2 > 0$.
    \item \label{c2p} $\alpha' = \beta' = 0$ and $\gamma' > 0$.
\end{enumerate}

\paragraph{First case: \ref{c1}\;+ \ref{c1p}}%
In the first case, we conclude that $v = \frac{1}{Q}$ and $f = \frac{P}{Q}$ where $Q$ is a polynomial of degree two, $Q > 0$ and $P$ is a polynomial of degree two or less.
Furthermore, $\lim_{-\infty} f = \lim_{+\infty} f = \frac{\alpha_0}{\alpha}$, regardless of whether $\alpha_0$ is zero or not. This implies that $f$ can't be invertible.
Going back to (\ref{modmodelequality}) and plugging these expressions:
\begin{multline}
\log p_x(x) + \half \log(\alpha x^2 + \beta x + \gamma)
            - \half \frac{\left(\alpha_0 x^2 + \beta_0 x + \gamma_0\right)^2}{\alpha x^2 + \beta x + \gamma}
            - \gamma_0'x + \half \gamma' x^2
            + (\alpha_0x^2 + \beta_0x)y - \half(\alpha x^2 + \beta x)y^2
\\=
\log p_y(y) + \half \log(\alpha' y^2 + \beta' y + \gamma')
            - \half \frac{\left(\alpha_0' y^2 + \beta_0' y + \gamma_0'\right)^2}{\alpha' y^2 + \beta' y + \gamma'}
            - \gamma_0y + \half \gamma y^2
            + (\alpha_0'y^2 + \beta_0'y)x - \half(\alpha' y^2 + \beta' y)x^2
\end{multline}
or again
\begin{equation}
\label{eq:long}
A(x) - B(y) -\half(\alpha - \alpha')x^2y^2 + \left(\alpha_0 - \half\beta'\right)x^2y - \left(\alpha_0' - \half\beta\right)xy^2 + (\beta_0 - \beta_0')xy = 0
\end{equation}
where
\begin{align}
    A(x) &= \log p_x(x) + \half \log(\alpha x^2 + \beta x + \gamma)
            - \half \frac{\left(\alpha_0 x^2 + \beta_0 x + \gamma_0\right)^2}{\alpha x^2 + \beta x + \gamma}
            - \gamma_0'x + \half \gamma' x^2 \\
    B(y) &= \log p_y(y) + \half \log(\alpha' y^2 + \beta' y + \gamma')
            - \half \frac{\left(\alpha_0' y^2 + \beta_0' y + \gamma_0'\right)^2}{\alpha' y^2 + \beta' y + \gamma'}
            - \gamma_0y + \half \gamma y^2
\end{align}
By first setting $x=0$ in equation~\eqref{eq:long}, we find that $A(x) = B(0)$. Similarly, by now setting $y=0$, we find that $B(y) = A(0)$.
This in particular means that $A(x) - B(y)$ is constant, which, when plugged back in equation~\eqref{eq:long}, would imply that all the  monomials are zero.
Finally, this would in turn imply the following:
\begin{gather}
    \alpha = \alpha', \; \alpha_0=-\half\beta', \; \alpha_0'=-\half\beta, \; \beta_0 = \beta_0' \\
    \log p_x(x) = - \half \gamma' x^2 + \gamma_0'x + \half \frac{\left(\alpha_0 x^2 + \beta_0 x + \gamma_0\right)^2}{\alpha x^2 + \beta x + \gamma} - \half \log(\alpha x^2 + \beta x + \gamma)  + C\\
    \log p_y(y) = - \half \gamma y^2 + \gamma_0y + \half \frac{\left(\alpha_0' y^2 + \beta_0' y + \gamma_0'\right)^2}{\alpha' y^2 + \beta' y + \gamma'} - \half \log(\alpha' y^2 + \beta' y + \gamma') + C
\end{gather}
Next we need to ensure we have well-defined probability densities.
From the above equations, we can check the coefficient of the quadratic term, which dominates at infinity, is $\frac{1}{2\alpha}(\alpha_0^2-\alpha\gamma')$ for $p_x$.  Requiring this to be negative is exactly the condition for the density family we made in Definition~\ref{def:lograt}.

For $p_y$, we get the dominant quadratic term with the coefficient $\frac{1}{2\alpha'}(\alpha_0'^2-\alpha'\gamma)$, and with substitutions we find the condition for its negativity as $\beta^2<4 \alpha \gamma$
which is, again, the same as a condition in the Definition.

Second, the constant $C$ has to be such that the probability density functions integrate to one. In fact, $C$ can be freely chosen, but importantly, it has to be the same for both densities. As a special case, this constraint is obviously fulfilled if the densities are the same, i.e.\ the parameters with and without prime are the same ($\alpha=\alpha'$ etc.). We shall show below that such parameter values can be found.

In fact, we can see how the parameters of the inverse model are determined from the parameters of the true model as follows. Define
\begin{equation}
    \delta:=\gamma' , \delta_0:=\gamma_0'
\end{equation}
So we can write the above as
\begin{gather}
     \log p_x(x) = - \half \delta x^2 + \delta_0 x + \half \frac{\left(\alpha_0 x^2 + \beta_0 x + \gamma_0\right)^2}{\alpha x^2 + \beta x + \gamma} - \half \log(\alpha x^2 + \beta x + \gamma)  + C\\
    \log p_y(y) = - \half \gamma y^2 + \gamma_0y + \half \frac{\left(-\beta y^2/2 + \beta_0 y + \delta_0\right)^2}{\alpha y^2 -2 \alpha_0 y + \delta} - \half \log(\alpha y^2 - 2 \alpha_0 y + \delta) + C
\end{gather}
where all the parameters defining $p_y$ are now obtained from the parameters defining $p_x,f,v$ (which are here denoted by the parameters without prime for this specific purpose). Likewise, we see that we also get $g$ and $w$ using those same parameters.

Now, we show that in spite of the different constraints, a solution in this family does exist. Let us consider the case where $p_x=p_y$, which would ensure that we can normalize the densities with a common $C$. This can be achieved by equating corresponding constants above which only requires
\begin{gather}
    \beta=-2\alpha_0\\
    \delta=\gamma\\
    \delta_0=\gamma_0
\end{gather}
which is still perfectly possible, even considering the constraints on the parameters in the Definition, which can be satisfied by simply taking non-negative $\alpha, \gamma, \gamma'$, and then fixing $\alpha_0$ to be small enough in absolute value (which implies the same for $\beta$). Thus, a solution for the inverse direction does exist. (But note we didn't prove that it exists for data coming from any $p_x,f,v$ in our family; we have proven unidentifiability  only for some parameter values.)

\paragraph{Second case: \ref{c2}\;+ \ref{c2p}}%
In the second case, we have that $v$ is constant.
Going back to (\ref{modmodelequality}), multiplying by $-2$, plugging the solutions just obtained:
\begin{equation}
- 2 \log p_x(x) + \gamma \left(y -\frac{\alpha_0}{\gamma} x^2 - \frac{\beta_0}{\gamma}x - \frac{\gamma_0}{\gamma}\right)^2 - \log \gamma
\\=
- 2 \log p_y(y) + \gamma' \left(x -\frac{\alpha_0'}{\gamma'} y^2 - \frac{\beta_0'}{\gamma'}y - \frac{\gamma_0'}{\gamma'}\right)^2 - \log \gamma'
\end{equation}
which can be expanded into, after grouping together monomials:
\begin{multline}
    - 2 \log p_x(x)
    + \frac{\alpha_0^2}{\gamma} x^4
    + 2\frac{\alpha_0\beta_0}{\gamma} x^3
    + \left(\frac{\beta_0^2 + \alpha_0\gamma_0}{\gamma} - \gamma'\right) x^2
    + 2\left(\frac{\beta_0\gamma_0}{\gamma} + \gamma_0'\right) x
    - 2\alpha_0 x^2y - 2\beta_0 xy
    + \textrm{const}
    \\=
    - 2 \log p_y(y)
    + \frac{{\alpha_0'}^2}{\gamma'} y^4
    + 2\frac{\alpha_0'\beta_0'}{\gamma'} y^3
    + \left(\frac{{\beta_0'}^2 + \alpha_0'\gamma_0'}{\gamma'} - \gamma\right) y^2
    + 2\left(\frac{\beta_0'\gamma_0'}{\gamma'} + \gamma_0\right) y
    - 2\alpha_0' y^2x - 2\beta_0'xy
\end{multline}
or again
\begin{equation}
\label{eq:postsimple}
A(x) - B(y) - 2\alpha_0x^2y + 2\alpha_0'y^2x + 2(\beta_0' - \beta_0)xy = \textrm{const}
\end{equation}
where
\begin{align}
\label{eq:xdist}
    A(x) &= - 2 \log p_x(x)
    + \frac{\alpha_0^2}{\gamma} x^4
    + 2\frac{\alpha_0\beta_0}{\gamma} x^3
    + \left(\frac{\beta_0^2 + \alpha_0\gamma_0}{\gamma} - \gamma'\right) x^2
    + 2\left(\frac{\beta_0\gamma_0}{\gamma} + \gamma_0'\right) x
\\
\label{eq:ydist}
    B(y) &= - 2 \log p_y(y)
    + \frac{{\alpha_0'}^2}{\gamma'} y^4
    + 2\frac{\alpha_0'\beta_0'}{\gamma'} y^3
    + \left(\frac{{\beta_0'}^2 + \alpha_0'\gamma_0'}{\gamma'} - \gamma\right) y^2
    + 2\left(\frac{\beta_0'\gamma_0'}{\gamma'} + \gamma_0\right) y
\end{align}

By setting $y=0$ in~\eqref{eq:postsimple}, we have that $A(x) = \textrm{const}$ for all $x$. Similarly, by setting $x=0$, we get $B(y) = \textrm{const}$ for all $y$.
We conclude that the remaining monomials must be zero.
In particular, this implies that $\alpha_0 = \alpha_0' = 0$ and $\beta_0 = \beta_0'$.
This in turn means that $f$ and $g$ are linear.

Finally, by plugging this into~\eqref{eq:xdist} and~\eqref{eq:ydist}, we get:
\begin{align}
\label{eq:xdist2}
    \log p_x(x) &= \half\left(\frac{\beta_0^2 }{\gamma} - \gamma'\right) x^2
    + \left(\frac{\beta_0\gamma_0}{\gamma} + \gamma_0'\right) x + \textrm{const}
\\
\label{eq:ydist2}
    \log p_y(y) &= \half\left(\frac{{\beta_0'}^2}{\gamma'} - \gamma\right) y^2
    + \left(\frac{\beta_0'\gamma_0'}{\gamma'} + \gamma_0\right) y + \textrm{const}'
\end{align}
We deduce that $x$ and $y$ must be Gaussian.
We don't prove the normalizability of the probability density functions in detail here since it is well-known that such Gaussian, unidentifiable models exist.

\paragraph{Third (and fourth) case: \ref{c1}\;+ \ref{c2p} or \ref{c2}\;+ \ref{c1p}}
Since these two cases are symmetric, we will suppose that $v$ is constant (\ref{c2}) and $\winv$ is a polynomial of second degree (\ref{c1p}).
Going back to \eqref{modmodelequality} and plugging the expressions for $f, v, g, w$:
\begin{multline}
\log p_y(y) + \half \log(\alpha' y^2 + \beta' y + \gamma')
            - \half \frac{\left(\alpha_0' y^2 + \beta_0' y + \gamma_0'\right)^2}{\alpha' y^2 + \beta' y + \gamma'}
            - \gamma_0y + \half \gamma y^2
            + (\alpha_0'y^2 + \beta_0'y)x - \half(\alpha' y^2 + \beta' y)x^2
        \\=
\log p_x(x) + \half \log(\gamma)
            - \half \frac{\left(\alpha_0 x^2 + \beta_0 x + \gamma_0\right)^2}{\gamma}
            - \gamma_0'x + \half \gamma' x^2
            + (\alpha_0x^2 + \beta_0x)y
\end{multline}

or again
\begin{equation}
\label{eq:long2}
A(x) - B(y) +\half\alpha'x^2y^2 + \left(\alpha_0 - \half\beta'\right)x^2y - \alpha_0'xy^2 + (\beta_0 - \beta_0')xy = 0
\end{equation}
where
\begin{align}
    A(x) &= \log p_x(x) + \half \log(\gamma)
            - \half \frac{\left(\alpha_0 x^2 + \beta_0 x + \gamma_0\right)^2}{\gamma}
            - \gamma_0'x + \half \gamma' x^2 \\
    B(y) &= \log p_y(y) + \half \log(\alpha' y^2 + \beta' y + \gamma')
            - \half \frac{\left(\alpha_0' y^2 + \beta_0' y + \gamma_0'\right)^2}{\alpha' y^2 + \beta' y + \gamma'}
            - \gamma_0y + \half \gamma y^2
\end{align}
Proceeding like above, we can deduce that $A(x) - B(y)$ is a constant, and that all the monomials in~\eqref{eq:long2} are zero. In particular, $\alpha' = 0$, which contradicts \ref{c1p}: this third case is thus not possible. \QED

\section{Affine flows are not universal density approximators}
\label{app:ununiversality}

\begin{proposition}
    Let $\Tb:\RR^d \rightarrow \RR^d$ be an affine autoregressive transformation. Let $\zb$ be a standard Gaussian, and let $\xb = \Tb(\zb)$. Then there is no parameterization of $\Tb$ such that $\xb$ has an isotropic Gumbel distribution.
\end{proposition}
\textit{Proof.}
It is enough to prove this Theorem for $d=2$.
Let $\xb = \Tb(\zb)$. Then
\begin{equation}
\label{eq:change}
    \log p_\xb(\xb) =\log p_\zb(\Tb^{-1}(\xb)) + \log \snorm{\det J_{\Tb^{-1}}(\xb)}
\end{equation}
and
\begin{align}
    x_1 &= e^{s_1} z_1 + t_1 \\
    x_2 &= e^{s_2(x_1)} z_2 + t_i(x_1)
\end{align}
The Jacobian log-determinant of $\Tb^{-1}$ is simply $\log\snorm{\det J_{\Tb^{-1}}(\zb)} = - s_1 - s_2(x_1)$.
Note that this determinant is only a function of $x_1$.
This is the main reason why affine autoregressive flows are not universal density approximators.

To see this, suppose that $x_1$ and $x_2$ are independent, and that each has a Gumbel distribution. Plugging this into equation~\eqref{eq:change}, we get
\begin{equation}
-\left(x_1 + e^{-x_1}\right) -\left(x_2 + e^{-x_2}\right) = -s_1 -s_2(x_1) - \left(x_1 - t_1\right)^2e^{-2s_1} - \left(x_2 - t_2(x_1)\right)^2e^{-2s_2(x_1)}
\end{equation}
This equation is valid for all $(x_1, x_2) \in \RR^2$.
In particular, let $x_1 = 0$. Then for any $x_2$, after rearranging and grouping terms, we get
\begin{equation}
    e^{-x_2} = \alpha x_2^2 + \beta x_2 + \gamma
\end{equation}
This can't hold for all values of $x_2$, which results in a contradiction.
Thus, we conclude that an affine autoregressive flow can't represent any distribution, unlike general unconstrained autoregressive flows. \QED

\section{Affine autoregressive flows are transitive}
\label{app:transitive}

\begin{proposition}
Consider $2$ autoregressive transformations $\fb$ and $\gb$ with the same ordering $\pi$.
Then their composition $\hb= \gb \circ \fb$ is also an autoregressive with the same ordering $\pi$.
\end{proposition}
\textit{Proof.}
Without loss of generality, assume that $\pi$ is the identity.
Let $(\xb, \yb, \zb)$ be such that
\begin{align}
    \yb &= \fb(\zb) \\
    \xb &= \gb(\yb) = \gb\circ\fb(\zb)
\end{align}
Since $\fb$ and $\gb$ are autoregressive, we can rewrite this system using equation~\eqref{eq:ar_flow} as
\begin{align}
y_i & = \tau(z_i, \yb_{<i}) \\
x_j & = \tau'(y_j, \xb_{<j})
\end{align}
The transformers $\tau$ and $\tau'$ are invertible with respect to their first argument. Denoting those inverses as $\alpha$ and $\alpha'$. Then
\begin{align}
z_i &= \alpha(y_i, y_{<i}) \\
y_j &= \alpha'(x_j, x_{<j})
\end{align}
And thus
\begin{equation}
\label{eq:zxar}
    z_i = \alpha(\alpha'(x_i, x_{<i}), \beta(x_{<i}))
\end{equation}
for some function $\beta$ (not necessarily invertible). Since $\alpha$ and $\alpha'$ are invertible with respect to their first argument, this means that the mapping $x_i \mapsto z_i$ in equation~\eqref{eq:zxar} is also invertible, and we can write:
\begin{equation}
    x_i = \tau''(z_i, x_{<i})
\end{equation}
where $\tau''$ is invertible wrt to its first argument.
This proves that $\hb = \gb\circ\fb$ is also an autoregressive flow. \QED

\begin{proposition}
\label{prop:transitive}
Consider $k$ affine autoregressive flows $\Tb_1, \dots, \Tb_k$ of the form~\eqref{eq:flow_def_x_ns} with the same ordering $\pi$.
Then their composition $\Tb=\Tb_1\circ \cdots \circ \Tb_k$ is also an affine autoregressive flow of the form~\eqref{eq:flow_def_x_ns} with the same ordering $\pi$.
\end{proposition}
\textit{Proof.}
We will suppose that $d=2$. The proof for $d>2$ is very similar but requires more complex notations.
We will denote by $z_l^j$ the $j$-th ($j=1,2$) output of the $l$-th sub-flow.
Not that we can parameterize $\Tb$ or $\Tb^{-1}$ to be an affine transformation.
In these notations, if $\Tb$ follows $\eqref{eq:flow_def_x_ns}$, then $\zb^k = \zb$ and $\zb^0 = \xb$.
If instead, $\Tb^{-1}$ follows $\eqref{eq:flow_def_x_ns}$, then $\zb^0 = \zb$ and $\zb^k = \xb$.
Each flow $l\geq1$ has the expression:
\begin{align}
z^l_1 &= \left(z^{l-1}_1 - t_1^{l}\right)e^{-s_1^{l}} \label{eq:flow_z1k}\\
z^l_2 &= \left(z^{l-1}_2 - t_2^{l}(z^l_1)\right)e^{-s_2^{l}(z^l_1)} \label{eq:flow_z2k}
\end{align}

First, define
\begin{align}
    \overline{s}_1^{l,k} &= \sum_{j=l+1}^ks_1^j \label{eq:s1lk}\\
    \overline{t}_1^{l,k} &= \sum_{j=l+1}^k t_1^je^{\sum_{i=l+1}^{j-1}s_1^i} = \sum_{j=l+1}^k t_1^je^{\overline{s}_1^{l,j-1}} \label{eq:t1lk}
\end{align}
where all sums are zero if they have no summands. Then it is easy to show by induction using~\eqref{eq:flow_z1k} that
\begin{equation}
\label{eq:z1lk}
z_1^l = e^{\overline{s}_1^{l,k}}z_1^k + \overline{t}_1^{l,k},\,\,\forall l\leq k
\end{equation}
and that
\begin{equation}
\label{eq:z1kk1}
e^{\overline{s}_1^{l,j}}z_1^j + \overline{t}_1^{l,j} = e^{\overline{s}_1^{l,k}}z_1^{k} + \overline{t}_1^{l,k}, \,\, \forall l\leq \min(j, k)
\end{equation}

Second, define
\begin{align}
    \overline{s}_2^k(u) &= \sum_{l=1}^k s_2^l( e^{\overline{s}_1^{l,k}}u + \overline{t}_1^{l,k} ) \\
    \overline{t}_2^k(u) &= \sum_{l=1}^k t_2^l( e^{\overline{s}_1^{l,k}}u + \overline{t}_1^{l,k} )
    e^{\sum_{i=1}^{l-1}s_2^i(e^{\overline{s}_1^{i,k}}u + \overline{t}_1^{i,k})}
\end{align}
We will show by induction on $k$ that
\begin{equation}
\label{eq:z2k}
    z^k_2 = \left(z_2^0 - \overline{t}^k_2(z_{1}^k)\right)e^{-\overline{s}^k_2(z_1^k)}
\end{equation}
The case for $k=1$ trivially holds. Now suppose that~\eqref{eq:z2k} holds for $k\geq1$, and let's show it also holds for $k+1$. Using~\eqref{eq:flow_z2k}, we can write
\begin{equation}
z_2^{k+1} = \left(z_2^k - t_2^{k+1}(z_1^{k+1}) \right)e^{-s_2^{k+1}(z_1^{k+1})}
\end{equation}
We need to show that
\begin{align}
    \overline{s}_2^{k+1}(z_1^{k+1}) &= s_2^{k+1}(z_1^{k+1}) + \overline{s}_2^k(z_1^k) \\
    \overline{t}_2^{k+1}(z_1^{k+1}) &= t_2^{k+1}(z_1^{k+1})e^{\overline{s}_2^k(z_1^k)} + \overline{t}_2^{k}(z_1^{k})
\end{align}
This can be done using~\eqref{eq:z1kk1}, the fact that $z_1^{k+1} = e^{\overline{s}_1^{k+1,k+1}}z_1^{k+1} + \overline{t}_1^{k+1,k+1}$ and the definitions of $\overline{s}_2^{k}$ and $\overline{t}_2^{k}$, which in turn allows us to conclude the induction proof.

Finally, by replacing $\zb^0$ and $\zb^k$ by $\xb$ and $\zb$ respectively, in~\eqref{eq:z1lk} and~\eqref{eq:z2k}, we have
\begin{align}
x_1 &= e^{\overline{s}_1^{0,k}}z_1 + \overline{t}_1^{0,k} \\
x_2 &= e^{\overline{s}^k_2(x_1)}z_2 + \overline{t}^k_2(x_1)
\end{align}
which proves the transitivity of affine autoregressive flows. \QED

\section{Universality of the causal function}
\label{app:universal}

\begin{proposition}
Consider $k$ affine autoregressive flows $\Tb_1, \dots, \Tb_k$, and let $\Tb=\Tb_1\circ \cdots \circ \Tb_k$.
Denote by $t_j^l$ and $s_j^l$ the coefficients of the $l$-th sub-flow $\Tb_l$, and by $\overline{t}_j^k$ and $\overline{s}_j^k$ those of $\Tb$.
Suppose that all of the $s_j^l$ and $t_j^l$ are feed-forward neural networks that have universal approximation capability (assuming all technical conditions hold).
Then  $\overline{t}_j^k$ and $\overline{s}_j^k$ also have universal approximation capability.
\end{proposition}

\textit{Proof.}
We will suppose for the proof that $d=2$. The proof for $d>2$ is similar.
According to Proposition~\ref{prop:transitive}, $\Tb$ is also an affine autoregressive flow, and $\overline{t}^k_2$ and $\overline{s}^k_2$ have the following expressions:
\begin{align}
    \overline{s}_2^k(u) &= \sum_{l=1}^k s_2^l( e^{\overline{s}_1^{l,k}}u + \overline{t}_1^{l,k} ) \label{eq:s2k2} \\
    \overline{t}_2^k(u) &= \sum_{l=1}^k t_2^l( e^{\overline{s}_1^{l,k}}u + \overline{t}_1^{l,k} )
    e^{\sum_{i=1}^{l-1}s_2^i(e^{\overline{s}_1^{i,k}}u + \overline{t}_1^{i,k})} \label{eq:t2k2}
\end{align}
where $\overline{t}_1^{l,k}$ and $\overline{s}_1^{l,k}$ are defined in equations~\eqref{eq:t1lk} and~\eqref{eq:s1lk} respectively.

On the one hand, translating and scaling the argument $u$ of $\overline{s}_2^k$ by $\overline{t}_1^{l,k}$ and $\overline{s}_1^{l,k}$ only changes the bias and the slope of the input layer of each of the $s_2^l$, $l=1,\dots,k$.
Thus, one can interpret equation~\eqref{eq:s2k2} as the output of an additional final layer of the neural network
whose outputs are the $s_2^l$ functions.
The number of flows $k$ in this case increases the width of this final layer.
Using the classical result of the universal approximation theorem of feed-forward networks with arbitrary width \citep{hornik1991approximation}, we conclude that $\overline{s}_2^k$ also satisfies such properties.

Interestingly, note that this results holds even if each of the $s_j^l$ function is simply an affine function followed by a nonlinearity (\ie a $1$-hidden layer feed-forward network).

On the other hand, since each of the $t_2^l$ have universal approximation capability, each can in particular approximate a function of the form $u\mapsto f_l(u)e^{\sum_{i=1}^{l-1}s_2^i(e^{\overline{s}_1^{i,k}}u + \overline{t}_1^{i,k})}$, where $f_l$ is a simple affine function followed by a nonlineariy $\sigma$ (\ie a $1$-hidden layer feed-forward network).
Thus, $\overline{t}_2^k$ can approximate a function of the form $\sum_{l=1}^k f_l$, which, by the same argument used above, will have universal approximation capability \citep{hornik1991approximation}. \QED

\section{Algorithms for causal inference}
\label{app:algorithms}

\subsection{Interventions}
\label{app:intervention}

As discussed in Section~\ref{sec:intervention}, the intervention $\textrm{do}(x_i=\alpha)$ breaks the links from $x_{<\pi(i)}$ to $x_i$ and sets a point mass on $z_i$.
Computing the value of $\xb_{j\neq i}$ requires sampling from $\prod_{j\neq i} p_{z_j}$ then propagating sequentially through the flow.
This avoids having to invert the flow and compute $z_i = \tau_i^{-1}(x_i, \xb_{<\pi(i)})$.
However, in the case of affine autoregressive flows, $tau^{-1}$ is readily available, and can be used to make the above algorithm parallelizable.
In fact, we can compute $z_i = \tau_i^{-1}(x_i, \xb_{<\pi(i)})$, sample $\zb_{j\neq i}$, then propagate the concatenated $\zb$ forward through the flow to obtain $\xb^{\textrm{do}(x_i=\alpha)}$.
Note that the value of $\xb_{<\pi(i)}$ is required to infer $z_i$, which will break the parallelism.
But since the same value is used to parametrized $\tau_i$ and $\tau_i^{-1}$, any value $\vb$ can be used as long as $\tau_i(\tau_i^{-1}(\alpha, \vb), \vb)=\alpha$.
In our implementation, we chose $\vb = 0$.
The sequential and parallel implementation are summarized by Algorithms~\ref{alg:internvention} and~\ref{alg:internvention_parallel} respectively.

\begin{algorithm}[ht]
    \caption{Generate samples from an interventional distribution (sequential)}
    \label{alg:internvention}
    \begin{algorithmic}
        \STATE {\bfseries Input:}  interventional variable $x_i$, intervention value $\alpha$, number of samples $S$
        \FOR{$s=1$ {\bfseries to} $S$}
            \STATE sample $\zb(s)$ from flow base distribution (the value of $z_i$ can be discarded)
            \STATE set $x_i(s) = \alpha$
            \FOR{$j = \pi^{-1}(1)$ to $\pi^{-1}(d)$; $j \neq i$}
                \STATE  compute observation $x_j(s) = \tau_j(z_j(s), \xb_{<\pi(j)}(s))$
            \ENDFOR
        \ENDFOR
        \STATE {\bfseries Return:} interventional sample $\mathbf{X} = \{\xb(s): s=1, \ldots, S\}$
    \end{algorithmic}
\end{algorithm}

\begin{algorithm}[ht]
    \caption{Generate samples from an interventional distribution (parallel)}
    \label{alg:internvention_parallel}
    \begin{algorithmic}
        \STATE {\bfseries Input:}  interventional variable $x_i$, intervention value $\alpha$, number of samples $S$
        \FOR{$s=1$ {\bfseries to} $S$}
            \STATE sample $\zb(s)$ from flow base distribution (the value of $z_i$ can be discarded)
            \STATE set $z_i (s)= \tau^{-1}\left(\alpha, \mathbf{0}\right)$
            \STATE compute $\xb(s) = \Tb(\zb(s))$
        \ENDFOR
        \STATE {\bfseries Return:} interventional sample $\mathbf{X} = \{\xb(s): s=1, \ldots, S\}$
    \end{algorithmic}
\end{algorithm}

\subsection{Counterfactuals}
\label{app:counterfactual}
The process of obtaining counterfactual predictions is described in
\citet{pearl2009causal} as consisting of three steps:
\begin{enumerate}
    \item \textbf{Abduction}: given an observation $\mathbf{x}^{obs}$, infer the conditional distribution/values over
    latent variables $\zb^{obs}$.
    In the context of an autoregressive flow model this is obtained as $\zb^{obs}= \Tb^{-1} ( \mathbf{x}^{obs} )$.
    \item \textbf{Action}: substitute
    the values of $\zb^{obs}$
    with the %
    values %
    based on the counterfactual query, $\mathbf{x}_{x_j \leftarrow \alpha}$.
    More concretely, for a counterfactual, $\mathbf{x}_{x_j \leftarrow \alpha}$, we
    replace the
    structural equations for $x_j$ with
    $x_j = \alpha$ and
    adjust the inferred value of latent $z^{obs}_j$ accordingly.

    \item \textbf{Prediction}:  compute the implied distribution over $\mathbf{x}$ by propagating latent variables, $\zb^{obs}$,
    through the structural equation models.
\end{enumerate}

This is summarized by Algorithm~\ref{alg:counterfactual}.

\begin{algorithm}[hb]
    \caption{Answer a counterfactual query}
    \label{alg:counterfactual}
    \begin{algorithmic}
        \STATE {\bfseries Input:}
        observed data $\textbf{x}^{obs}$,
        counterfactual variable $x_j$ and value $\alpha$

        \STATE {\bfseries \quad 1. Abduction}: infer $\textbf{z}^{obs} = \Tb^{-1} ( \mathbf{x}^{obs})$
        \STATE {\bfseries \quad 2. Action}: $(a)$ set $z_{j, x_j\leftarrow \alpha}^{obs} = \tau_j^{-1}(\alpha, \xb_{<\pi(j)}^{obs})$
        \STATE \hspace{2.215cm} $(b)$ set $z_{i, x_j\leftarrow \alpha}^{obs} = z_i^{obs}$ for $i \neq j$
        \STATE {\bfseries \quad 3. Prediction}: pass $\zb^{obs}_{x_j\leftarrow \alpha}$ forward through the flow $\Tb$

        {\bfseries Return:} $\mathbf{x}_{x_j \leftarrow \alpha} = \Tb(\zb^{obs}_{x_j\leftarrow \alpha})$
    \end{algorithmic}
\end{algorithm}

\section{Experimental details}
\label{app:exp}

\subsection{Architectures and hyperparameters}
\label{app:arch}

The optimization was done using Adam, with learning rate $\verb+lr+=0.001$, $\betab = (0.9, 0.999)$, along with a scheduler that reduces the learning rate by a factor of $0.1$ on plateaux.
All flows use an isotropic Laplace distribution as a prior.
The different architectures and hyperparameter used for the experiments are as follows:
\begin{itemize}
    \item \textbf{Causal discovery simulations:} The flow $\Tb$ is a composition of $2$ sub-flows $\Tb_1$ and $\Tb_2$. For each of the $\Tb_l$, both $s_j$ and $t_j$ are multi-layer perceptrons (MLPs), with $1$ hidden layer and $10$ hidden units. Each direction was trained for $200$ epochs, with a mini-batch of $128$ data points.
    The same architecture was used for all panels of Figure~\ref{Fig:causalDiscSimulations}.
    \item \textbf{Cause-effect pairs:} The flow $\Tb$ is a composition of $4$ sub-flows $\Tb_1, \cdots, \Tb_4$. For each of the $\Tb_l$, both $s_j$ and $t_j$ are MLPs, with either $1$ or $3$ hidden layers, each with $5$ hidden units.
    For each direction, we train two different flows (with $1$ or $3$ hidden layers), and select the flow that yields higher test likelihood.
    Each direction was trained for $750$ epochs, with a mini-batch of $128$ data points.
    For each pair, $80\%$ of the data points were used for training, and the remaining $20\%$ to evaluate the likelihood.
    The same architecture was used to classify all the pairs.
    \item \textbf{EEG arrow of time:} The flow $\Tb$ is a composition of $4$ sub-flows $\Tb_1, \cdots, \Tb_4$. For each of the $\Tb_l$, both $s_j$ and $t_j$ are MLPs, with $4$ hidden layers, each with $10$ hidden units.
    Each direction was trained for $400$ epochs, with a mini-batch of $32$ data points.
    For each channel, $80\%$ of the data points were used for training, and the remaining $20\%$ to evaluate the likelihood.
    The same architecture was used to classify all the channels.
    \item \textbf{Interventions on simulated data:} The flow $\Tb$ is a composition of $5$ sub-flows $\Tb_1, \cdots, \Tb_5$.
    For each of the $\Tb_l$, both $s_j$ and $t_j$ are MLPs, with $1$ hidden layers, each with $10$ hidden units.
    We train the flow, conditioned on the causal ordering, to fit the correct SEM.
    Training was done for $750$ epochs, with a mini-batch of $32$ data points.
    \item \textbf{Interventions on es-fMRI data:}
    The flow $\Tb$ is a composition of $5$ sub-flows $\Tb_1, \cdots, \Tb_5$. For each of the $\Tb_l$, both $s_j$ and $t_j$ are MLPs, with a single hidden layer consisting of $2$ hidden units.
    In order to obtain interventional predictions, a \carefl model was
    first trained using resting-state fMRI data conditioned upon the causal ordering. Since we did not seek to infer the causal structure, 100$\%$ of the training data was employed (this is in contrast to causal discovery experiments which only trained models on $80\%$ of the data).
\end{itemize}

\subsection{Exploring flow architectures}
\begin{figure}
        \centering
        \includegraphics[width=.98\textwidth]{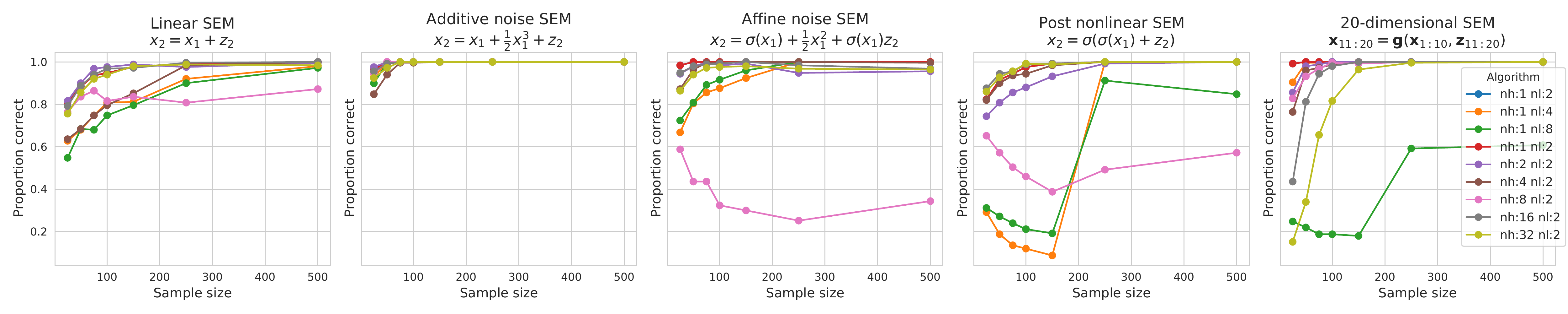}
        \caption{Impact of changing the width versus the depth of the normalizing flow in \carefle}
        \label{fig:widthdepth}
\end{figure}

As discussed in Section~\ref{app:transitive}, stacking multiple autoregressive flows on top of each other is equivalent to using a single autoregressive flow with a wide hidden layer.
To explore this aspect, we run multiple experiments where each flow is an MLP with one hidden layer and a LeakyReLU activation, in which we vary the width of the hidden layer and the number of stacked flows.
We observed empirically that stacking multiple layers in the flows lead to empirical improvements, as reported by Figure~\ref{fig:widthdepth}.

\subsection{Preprocessing of EEG data}
\label{app:eeg}
The openly available EEG data from \citet{dornhege2004boosting} contains recordings for 5 healthy subjects.
For each subject, the data has been sampled at 100Mhz and 1000Mhz.
For our experiments, we considered subject number 3, and used the data sampled at 1000Mhz.
In particular, we only considered $n=150$ and $n=500$ time points.
Each of the 118 EEG channels was then reversed with probability $0.5$.

The task is to properly infer the arrow of time for each of the 118 EEG, considered separately.
We transform a univariate timeseries $(x_t)_{t\in\iset{1,n}}$ corresponding to 1 channel into bivariate causal data by shifting it by a lag parameter $l$, to obtain data of the form $(x_t, x_{t+l})_{t\in\iset{1,n-l}}$.
For the results plotted in Figure~\ref{Fig:eeg}, we used three values of lag for ANM, RECI, the linear LR and \carefle-NS: $l\in\{1, 2, 3\}$, which we then combined into one dataset.
For \carefle, we used only two values of lag: $l\in\{1, 2\}$.

\subsection{Preprocessing of functional MRI data}
\label{app:esfmri}

Results included in this manuscript come from preprocessing performed using \verb+FMRIPREP+ \citep{esteban2019fmriprep}, a \verb+Nipype+ based tool \citep{gorgolewski2011nipype}. Each T1w (T1-weighted) volume was corrected for INU (intensity non-uniformity) using \verb+N4BiasFieldCorrection v2.1.0+ and skull-stripped using \verb+antsBrainExtraction.sh v2.1.0+ (using the OASIS template). Brain surfaces were reconstructed using recon-all from \verb+FreeSurfer v6.0.1+, and the brain mask estimated previously was refined with a custom variation of the method to reconcile \verb+ANTs+-derived and \verb+FreeSurfer+-derived segmentations of the cortical grey-matter of Mindboggle. Spatial normalization to the ICBM 152 Non-linear Asymmetrical template version 2009c  was performed through non-linear registration with the \verb+antsRegistration+ tool of \verb+ANTs v2.1.0+, using brain-extracted versions of both T1w volume and template. Brain tissue segmentation of cerebrospinal fluid (CSF), white-matter (WM) and gray-matter (GM) was performed on the brain-extracted T1w using \verb+fast+.

Functional data was slice time corrected using \verb+3dTshift+ from \verb+AFNI v16.2.07+ and motion corrected using \verb+mcflirt+. This was followed by co-registration to the corresponding T1w using boundary-based registration with six degrees of freedom, using \verb+bbregister+ (\verb+FreeSurfer v6.0.1+). Motion correcting transformations, BOLD-to-T1w transformation and T1w-to-template (MNI) warp were concatenated and applied in a single step using \verb+antsApplyTransforms+ using Lanczos interpolation.

Regional time series were subsequently calculated from the processed FMRI data (transformed into MNI space) using \verb+NiLearn+ \citep{abraham2014machine} and the Harvard-Atlas probabilistic atlas, with regions thresholded at 25$\%$ probability and binarised. Given the regional location of intracortical stimulation in the subjects, FMRI time-series from the  Cingulate gyrus and Heschl’s gyrus were selected for analysis.

We note that each patient
received surgery and stimulation in different locations, as determined by their diagnosis and clinical criteria. As such,
the two regions studied were selected so as to include as
many subjects as possible in our experiments.
Moreover, the Cingulate gyrus is a region associated with cognitive processes such as saliency and emotional processing \citep{vogt2019cingulate} whereas Heschl’s gyrus covers primary auditory cortex, associated with early cortical processing of auditory information; as such connectivity between the regions captures the interaction between a higher-order heteromodal region and a unimodal sensoryl region.

\end{document}